\documentclass{article}

\usepackage{microtype}
\usepackage{graphicx}
\usepackage{subfigure}
\usepackage{booktabs} 

\usepackage{hyperref}
\usepackage{multicol}
\usepackage{multirow}
\usepackage{comment}
\usepackage{bm}
\usepackage{enumitem}
\usepackage{pifont}
\usepackage{bbding}
\usepackage[framemethod=TikZ]{mdframed}

\usepackage[ruled,vlined]{algorithm2e}
\usepackage{tcolorbox}

\usepackage{xcolor}
\usepackage{color}
\usepackage{colortbl}
\tcbuselibrary{skins}
\usepackage{hyperref}  
\hypersetup{  
    colorlinks=true, 
    linkcolor=blue,
    urlcolor=red  
}

\newenvironment{qbox}
{\begin{tcolorbox}[enhanced jigsaw, drop shadow=black!50!white,colback=white, width=0.95\linewidth, center, left=2pt,right=2pt,top=1pt,bottom=1pt]}
{\end{tcolorbox}}

\usepackage[accepted]{icml2024}

\usepackage{amsmath}
\usepackage{amssymb}
\usepackage{mathtools}
\usepackage{amsthm}
\usepackage{lipsum}
\usepackage{titletoc}
\usepackage{minitoc}
\usepackage[page,header]{appendix}

\usepackage[capitalize,noabbrev]{cleveref}

\theoremstyle{plain}
\newtheorem{theorem}{Theorem}[section]

\newtheorem{corollary}[theorem]{Corollary}
\theoremstyle{definition}

\newcommand{\name}{ReconBoost}
\usepackage[textsize=tiny]{todonotes}
\definecolor{myblue}{RGB}{235, 244, 246}
\definecolor{mygreen}{RGB}{234, 240, 225}

\newtheorem*{rthm1}{\textbf{Restate of Theorem} \ref{thm:main}}
\newtheorem*{gthm1}{\textbf{Restate of Corollary} \ref{cor:main}}

\icmltitlerunning{ReconBoost: Boosting Can Achieve Modality Reconcilement}

\begin{document}

\twocolumn[
\icmltitle{ReconBoost: Boosting Can Achieve Modality Reconcilement}




\begin{icmlauthorlist}
\icmlauthor{Cong Hua}{ict,ucas-cs}
\icmlauthor{Qianqian Xu}{ict}
\icmlauthor{Shilong Bao}{iie,ucas-scs}
\icmlauthor{Zhiyong Yang}{ucas-cs}
\icmlauthor{Qingming Huang}{ucas-cs,ict,bdkm}
\end{icmlauthorlist}

\icmlaffiliation{ict}{Key Laboratory of Intelligent Information Processing, Institute of Computing Technology, Chinese Academy of Sciences, Beijing, China}
\icmlaffiliation{ucas-cs}{School of Computer Science and Technology, University of Chinese Academy of Sciences, Beijing, China}
\icmlaffiliation{iie}{Institute of Information Engineering, Chinese Academy of Sciences, Beijing, China}
\icmlaffiliation{ucas-scs}{School of Cyber Security, University of Chinese Academy of Sciences, Beijing, China}
\icmlaffiliation{bdkm}{Key Laboratory of Big Data Mining and Knowledge Management, Chinese Academy of Sciences, Beijing, China}

\icmlcorrespondingauthor{Qianqian Xu}{xuqianqian@ict.ac.cn}
\icmlcorrespondingauthor{Qingming Huang}{qmhuang@ucas.ac.cn}

\icmlkeywords{Multi-modal learning, modality, boosting,ICML}

\vskip 0.2in
]



\printAffiliationsAndNotice{}  

\begin{abstract}
This paper explores a novel multi-modal \textit{alternating} learning paradigm pursuing a reconciliation between the exploitation of uni-modal features and the exploration of cross-modal interactions. This is motivated by the fact that current paradigms of multi-modal learning tend to explore multi-modal features simultaneously. The resulting gradient prohibits further exploitation of the features in the weak modality, leading to modality competition, where the dominant modality overpowers the learning process. To address this issue, we study the modality-alternating learning paradigm to achieve reconcilement. Specifically, we propose a new method called \textit{ReconBoost} to update a fixed modality each time. Herein, the learning objective is dynamically adjusted with a reconcilement regularization against competition with the historical models. By choosing a KL-based reconcilement, we show that the proposed method resembles Friedman's Gradient-Boosting (GB) algorithm, where the updated learner can correct errors made by others and help enhance the overall performance. The major difference with the classic GB is that we only preserve the newest model for each modality to avoid overfitting caused by ensembling strong learners. Furthermore, we propose a memory consolidation scheme and a global rectification scheme to make this strategy more effective. Experiments over six multi-modal benchmarks speak to the efficacy of the method. We release the code at \url{https://github.com/huacong/ReconBoost}.
\end{abstract}

\section{Introduction}

Deep learning has significantly advanced uni-modal tasks \cite{resnet,jinhui-pami-17,jinhui-pami-19,jinhui-nips-20}. However, most real-world data usually follows a multi-modal nature (say text, video, and audio) in various fields such as data mining \cite{jiangdm2c_5,jiangdm2c}, computer vision \cite{event_camera,optical_flow,shao2024joint}, and medical diagnosis \cite{medical_deep_learning,hypergraph}. Because of this, the deep learning community has recently focused more on multi-modal learning \cite{,wei2022-ml-survey,hsr,shrec,zhang-ml-survey}.
The prevailing paradigm in multi-modal learning typically employs a \textbf{joint learning} strategy, wherein a wealth of studies \cite{fusion_1,fusion_2,fusion_3,fusion_4,dynamic_fusion,yuehang-tmm,shao2023identity} primarily focus on integrating modality-specific features into a \textbf{shared representation} for various downstream tasks.

 \begin{figure}[!t]  
\centering  
\begin{minipage}[t]{0.99\linewidth}  
\centering  
\subfigure[Audio Modality]{  
\includegraphics[width=0.30\linewidth]{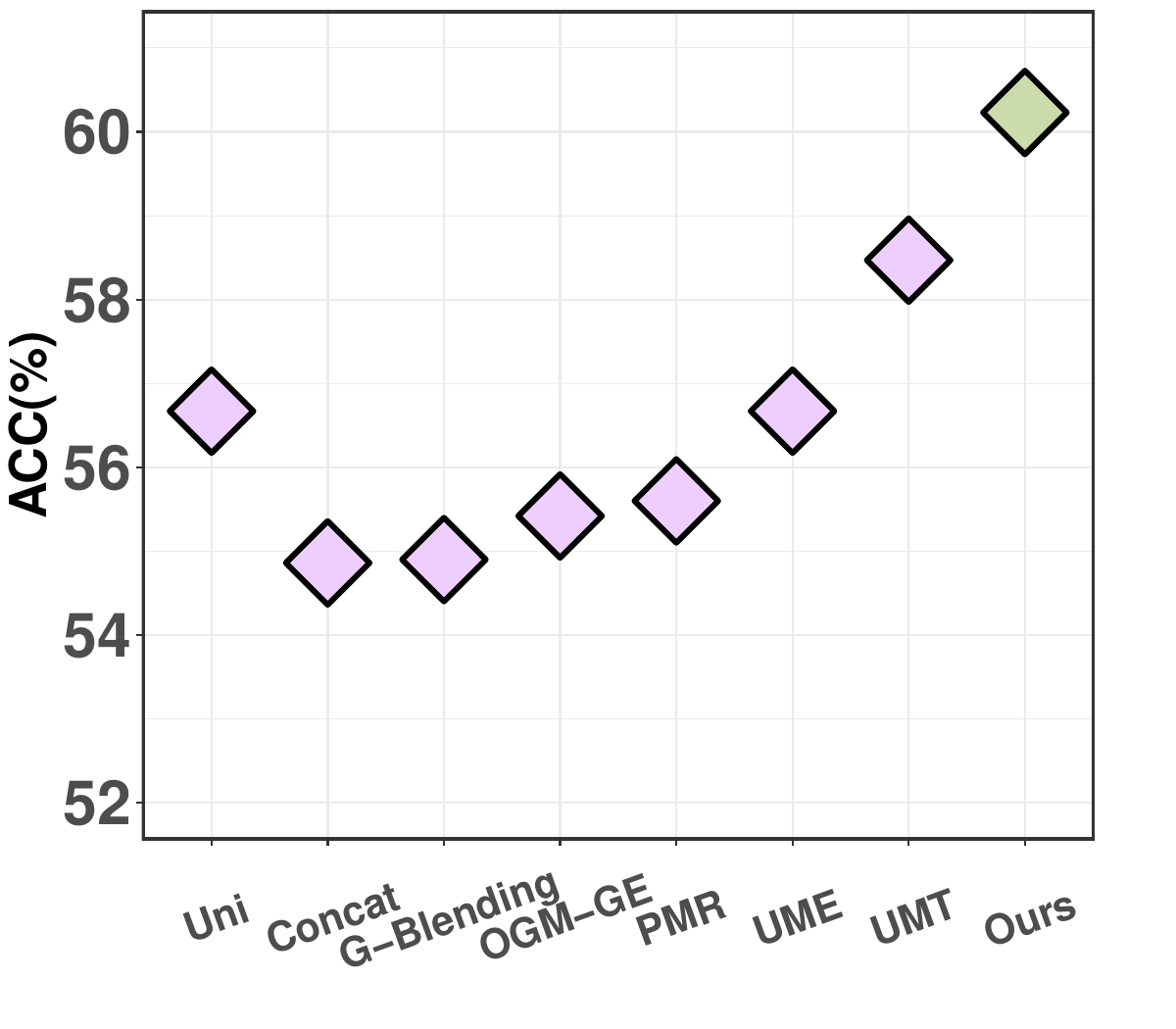} 
}  
\subfigure[Visual Modality]{  
\includegraphics[width=0.30\linewidth]{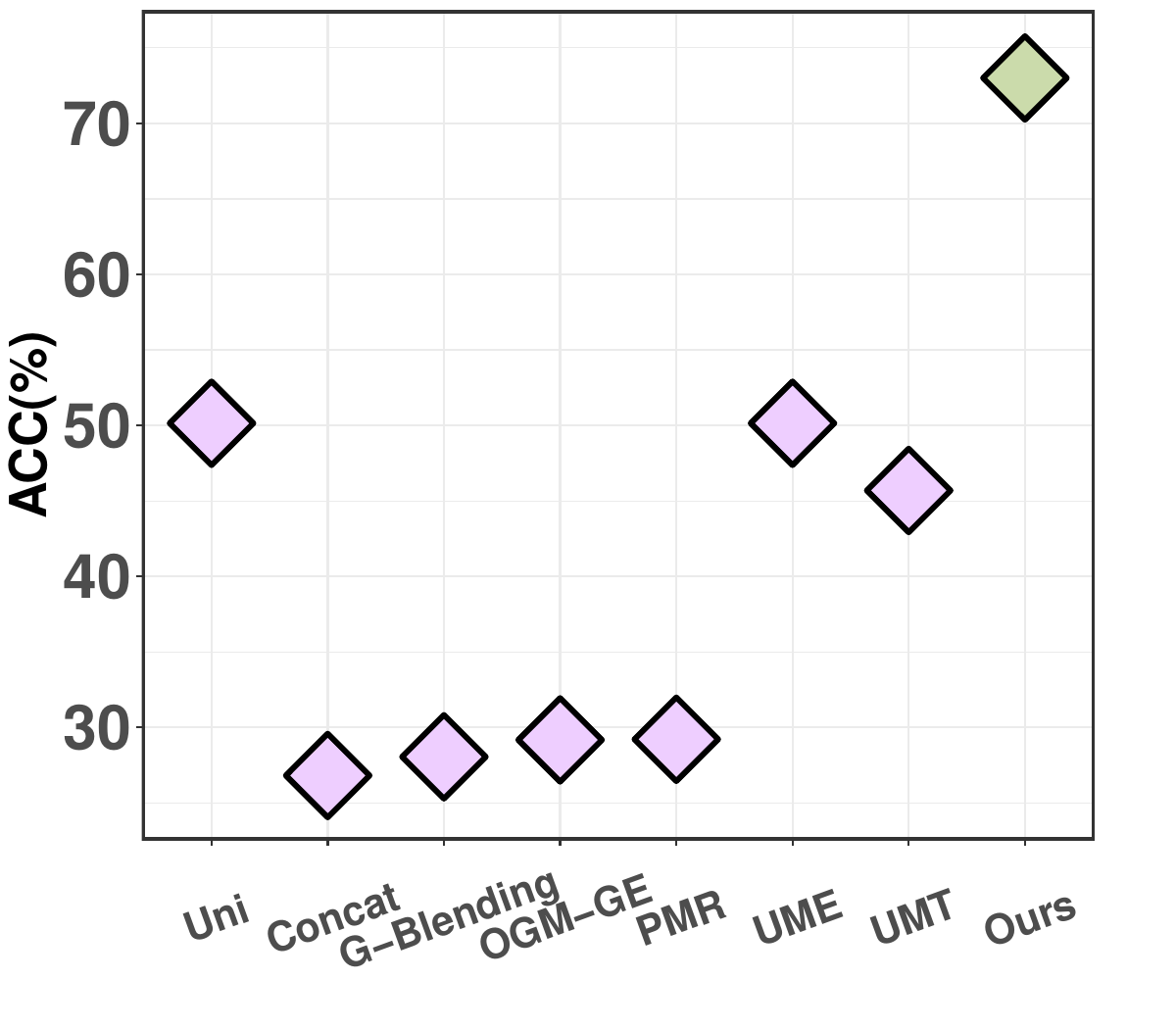}  
}  
\subfigure[Multi-modal]{  
\includegraphics[width=0.30\linewidth]{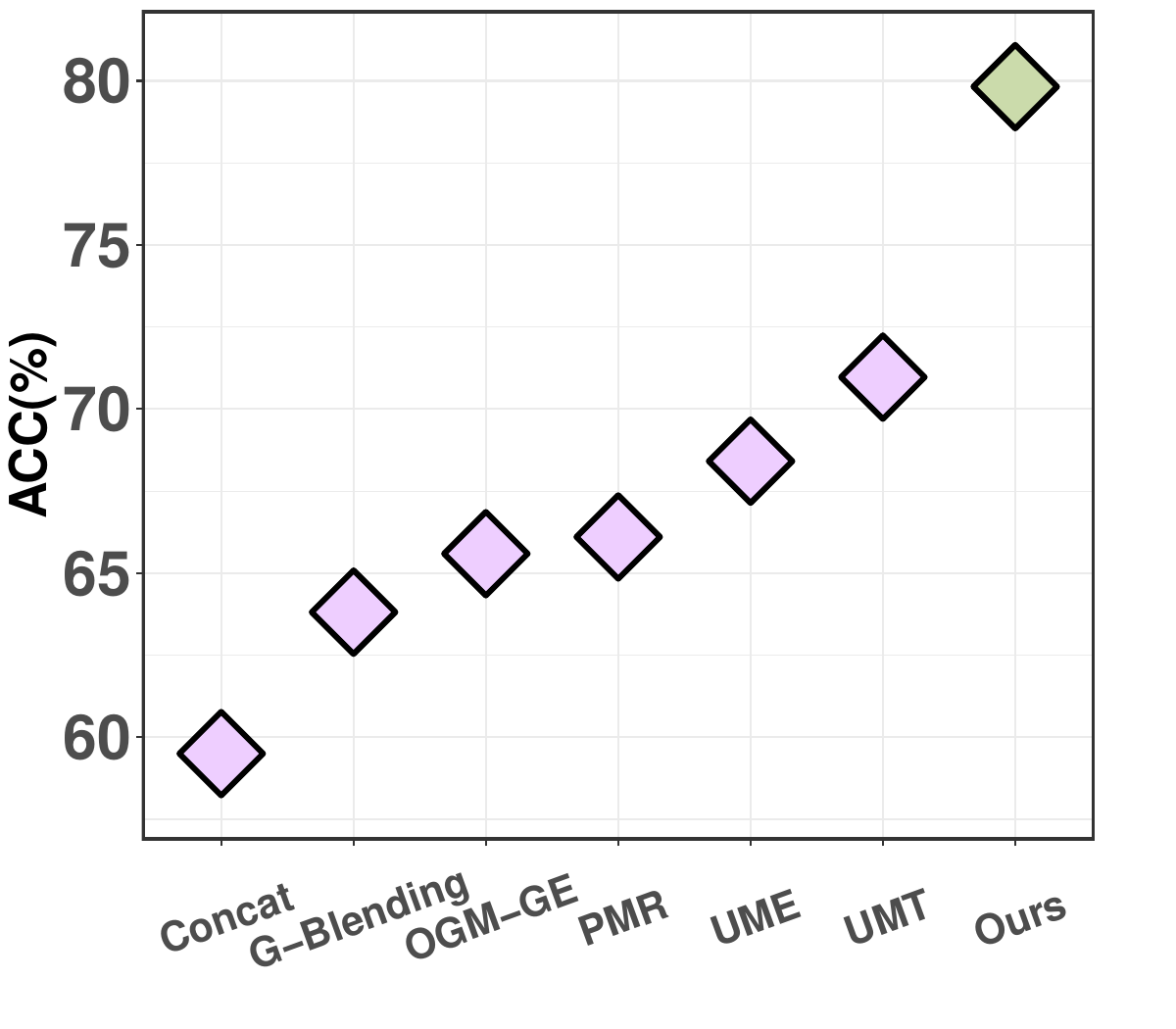}  
}  
\caption{The performance among multi-modal learning competitors on the CREMA-D dataset. For audio modality and visual modality, we evaluate the encoders of different competitors by training linear classifiers on them. Uni represents the uni-modal training method.}  
\label{fig:ml}
\end{minipage}  
\end{figure}

Despite great success, numerous experimental observations \cite{UMT,OGM_GE_2022_CVPR} and recent theoretical evidence \cite{modality_competition} have pointed out that current paradigms of multi-modal learning encounter \textbf{\textit{Modality Competition,}} where the model is dominated by some of the modalities. Various studies \cite{OGM_GE_2022_CVPR,PMR_CVPR_23,UMT} have been made to mitigate the modality competition issue. The primary concern is how to balance optimization progress across multi-modal learners and improve uni-modal feature exploitations. For instance, G-Blending \cite{g-blending} adds a uni-modal classifier with additional supervised signals in multi-modal learning to blend modalities effectively. OGM \cite{OGM_GE_2022_CVPR} and PMR \cite{PMR_CVPR_23} work on reducing the influence of the dominant modality and aiding the training of others through adaptive gradient modulation and dynamic loss functions, respectively. Besides, UMT \cite{UMT} distills knowledge from well-trained uni-modal models in multi-modal learning which can effectively benefit from uni-modal learning.

However, as depicted in Fig.\ref{fig:ml}, existing algorithms \textbf{still follow the joint learning strategy}, suffering from limited performance trade-offs for modality competition. Expanding the gradient update rule, we find that joint learning tends to neglect the gradient from weak modality. The dominant modality that converges more quickly would eventually overpower the whole learning process. 

Therefore, in this paper, we turn to the following question:
\begin{qbox}
\begin{center}
\textit{Can we achieve modality reconciliation via other learning paradigms? }
\end{center}
\end{qbox}

In search of an answer, we propose an effective method named \textbf{\textit{ReconBoost}}, where we alternate the learning for each modality. Intuitively, it naturally alleviates modality competition in the gradient space since the modality-specific gradients must be employed separately. To further enhance the effect of individual modalities, we propose a reconcilement regularization to maximize the diversity between the current update and historical models. Dynamically adjusting the learning objective via the regularization term further alleviates the modality competition issue induced by sticking to a particular modality. Theoretically, we show that by choosing a KL divergence \cite{kl-divergence} based reconcilement term, our proposed method can realize an alternating version of the well-known gradient boosting method  \cite{gb}. Specifically, the updated modality learner can focus on the errors made by others, thereby highlighting their complementarity. Unlike traditional boosting techniques \cite{freund1995boosting,freund1997decision}, which use weak learners like decision trees, our method employs DNN-based learners which are over-parameterized models. To avoid overfitting, we discard historical learners and only preserve the last learner for each modality, creating an \textbf{\textit{alternating-boosting strategy.}}  Additionally, considering the differences between the traditional boosting techniques and our alternating-boosting strategy, we present a memory consolidation scheme and a global rectification scheme to reduce the risk of forgetting critical patterns in historical updates.

Finally, we conduct empirical experiments on six multi-modal benchmarks and demonstrate that 1) ReconBoost can consistently outperform all the competitors significantly on all datasets. 2) ReconBoost can achieve \textbf{\textit{Modality Reconcilement.}}

\section{Preliminary}

In this section, we first review the task of multi-modal learning. Then, we further explain the current difficulties encountered in multi-modal learning. \textbf{\textit{Due to space limitations, we present a brief overview of prior arts in App.\ref{sec:related}.}}

\subsection{The Task of Multi-modal Learning}

Let $\mathcal{D}_{train} = \left\{(x_i,y_i)\right\}_{i=1}^{N}$  be a multi-modal dataset, where $N$ is the number of examples in the training set. Herein, each example $i$ consists of a set of raw features $x_i = \{m_i^k\}_{k=1}^{M}$ from different $M$ modalities and a one-hot label $y_i = \{c_{i,j}\}_{j=1}^{Y}$ ,  where $c_{i,j} = 1$ if the label for $i$ is j, otherwise  $c_{i,j} = 0$; $Y$ is the total number of categories.

Given $M$ modality-specific feature extractors $\{\mathcal{F}_{k}(\theta_k)\}_{k=1}^{M}$, with $\mathcal{F}_{k}$ typically being a deep neural network with parameters $\theta_k$, $\{\mathcal{F}_k(\theta_k;m_i^{k})\}_{k=1}^{M}$ denotes the latent features of $i$-th sample, where $\mathcal{F}_k(\theta_k;m_i^{k}) \in \mathbb{R}^{d_k}$. Then, we define the predictor $\mathcal{S}$ as a mapping from the latent feature space to the label space. The \textbf{objective} of multi-modal learning is to jointly minimize the empirical loss of the predictor:
\begin{equation}
    \small
    \begin{split}
         \mathcal{L}(\mathcal{S}(\{\mathcal{F}_k(x)\}_{k=1}^{M}),y) = \frac{1}{N}\sum_{i=1}^{N}\ell(\mathcal{S}(\{\mathcal{F}_k(\theta_k;m_i^{k})\}_{k=1}^{M}),y_i) 
    \end{split}
\end{equation}
where $\ell$ is the CE loss. In multi-modal learning, a key step is to merge the modality-specific representations. To this end, the predictor is often formalized as a composition: $ g \circ f $, where $g$ is a simple classifier and $f$ is a cross-modal fusion strategy.
For example, one can use the concatenate operation to implement the fusion strategy and use a linear model to implement the classifier. In this case, the resulting predictor becomes:
\begin{equation}\small
\begin{aligned}
    \small
    \begin{split}
    \mathcal{S}(\{\mathcal{F}_k(\theta_k;m_i^{k})\}_{k=1}^{M}) & = W\cdot[\mathcal{F}_1(\theta_1;m_i^1):\cdots:\mathcal{F}_M(\theta_k;m_i^k)] \\
    & = \sum_{k} W_k \cdot \mathcal{F}_k(\theta_k;m_i^k),
    \end{split}
\end{aligned}    
\end{equation}
where $W \in \mathbb{R}^{Y\times \sum_{k}d_k}$ is the last linear classifier, $W_k \in {\mathbb{R}^{Y\times d_k}}$ is a block of $W$ for the $k$-th modality. 

In contrast to uni-modal training, information fusion in multi-modal learning can help explore cross-modal interactions, enhancing performance across various real-world scenarios. However, under the current paradigm of multi-modal learning, the limitations in effectively exploiting uni-modal features have constrained the performance of the multi-modal learning model. We state the corresponding challenges herein in the upcoming subsection.
\begin{figure}[]
\centering  
\subfigure[Testing Acc. on CREMA-D]{
\includegraphics[width=0.48\columnwidth]{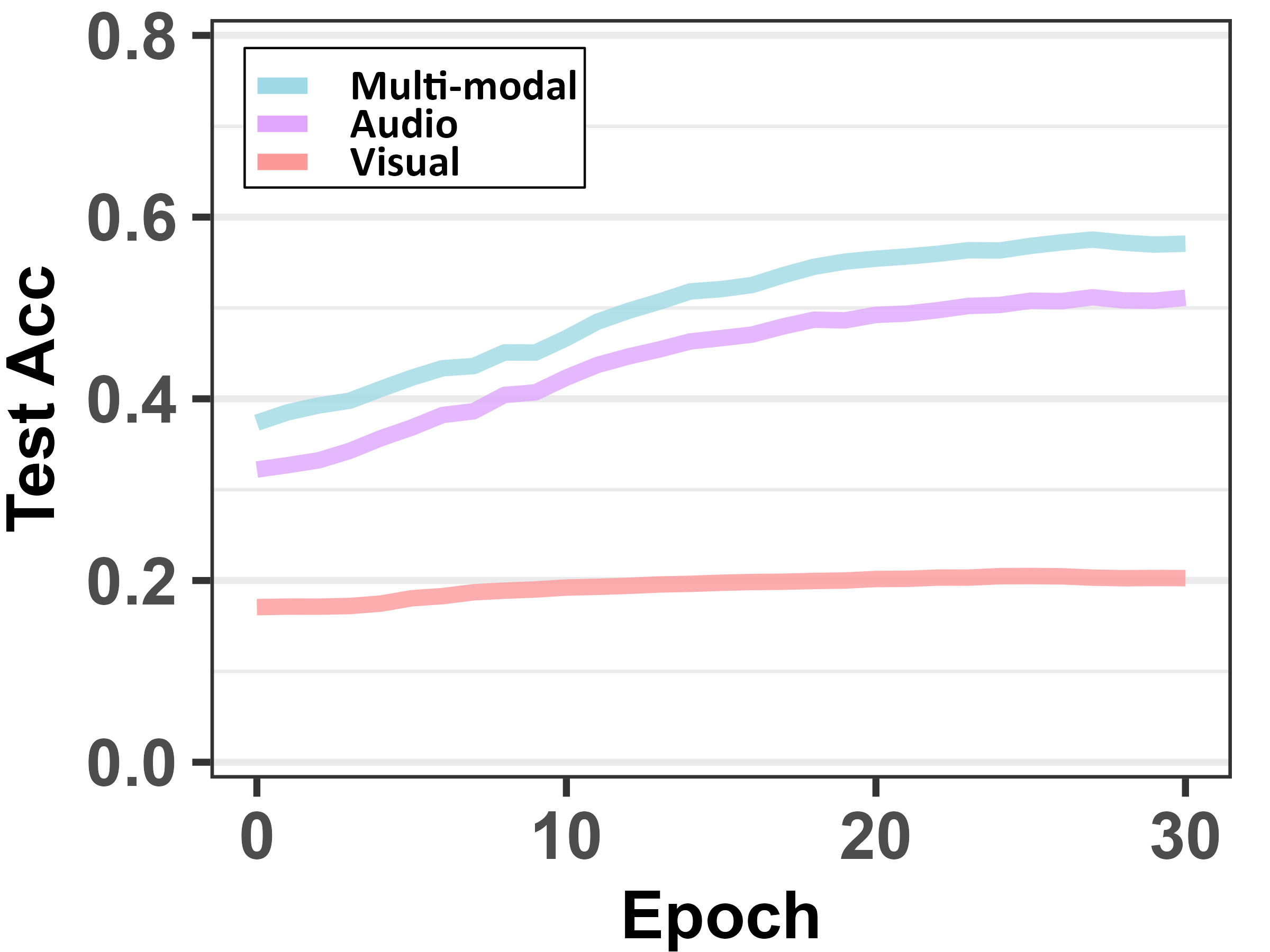}}
\subfigure[Training Loss on CREMA-D]{
\includegraphics[width=0.48\columnwidth]{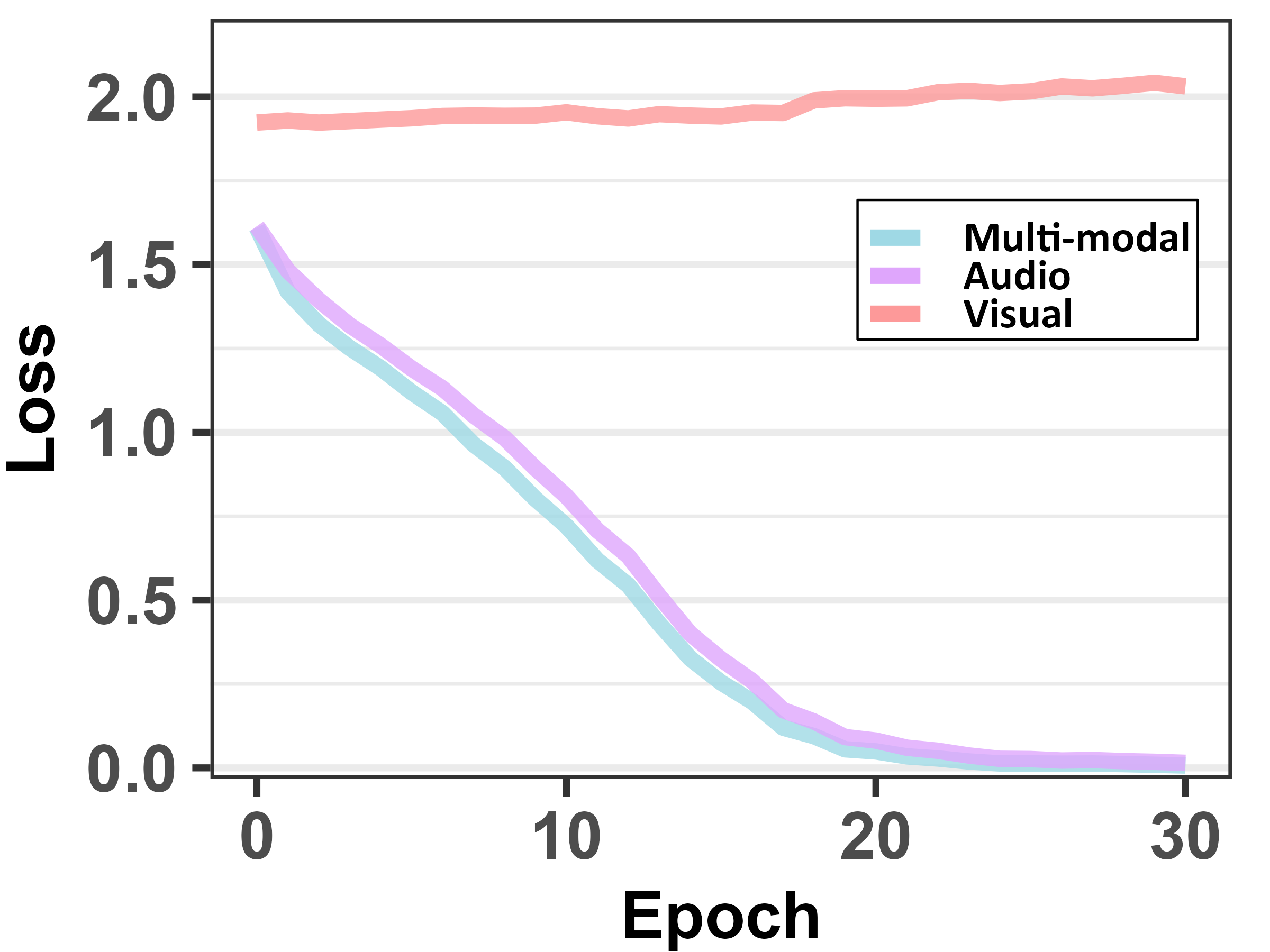}}
\subfigure[Testing Acc. on MOSEI]{
\includegraphics[width=0.48\columnwidth]{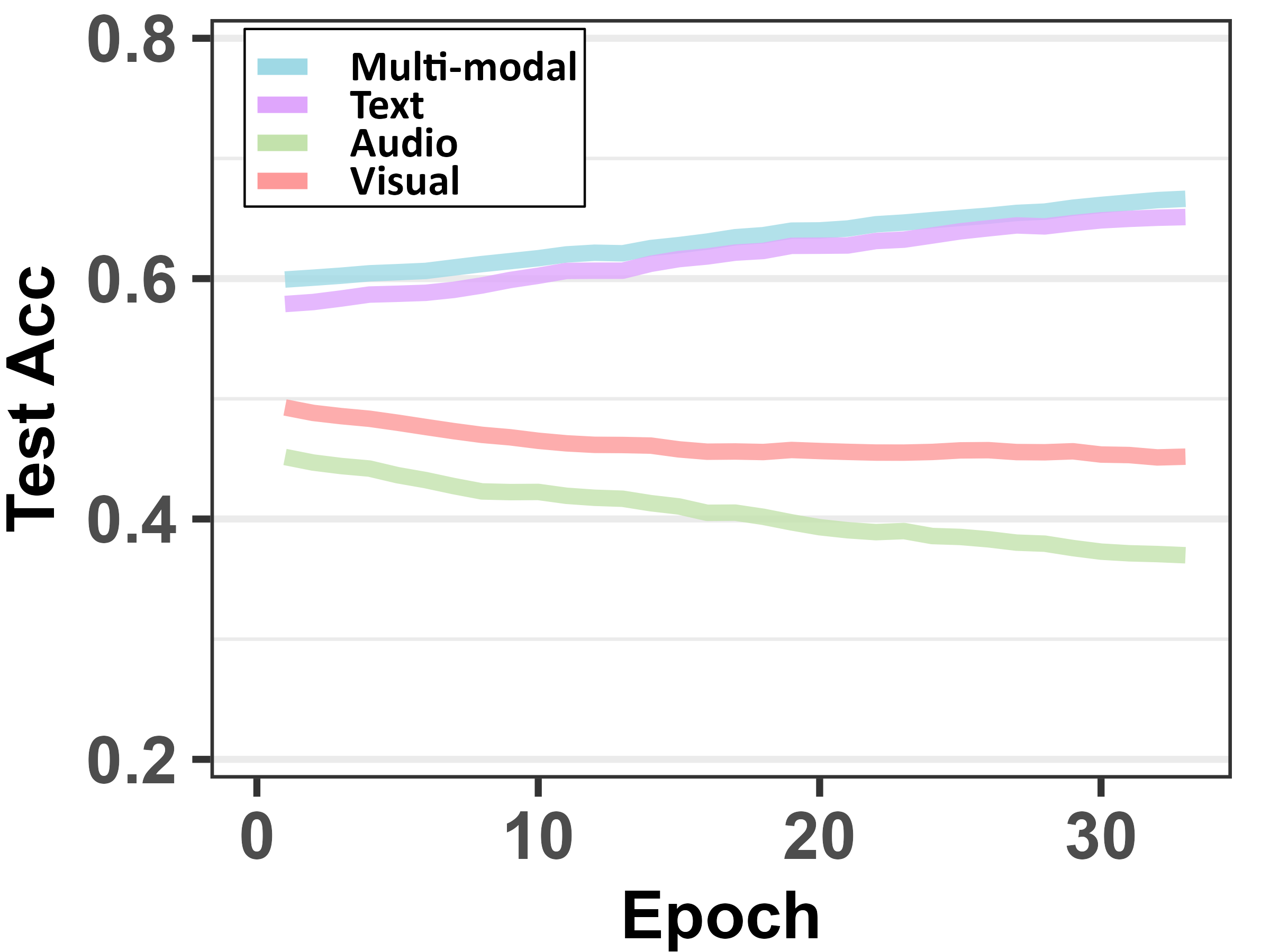}}
\subfigure[Training Loss on MOSEI]{
\includegraphics[width=0.48\columnwidth]{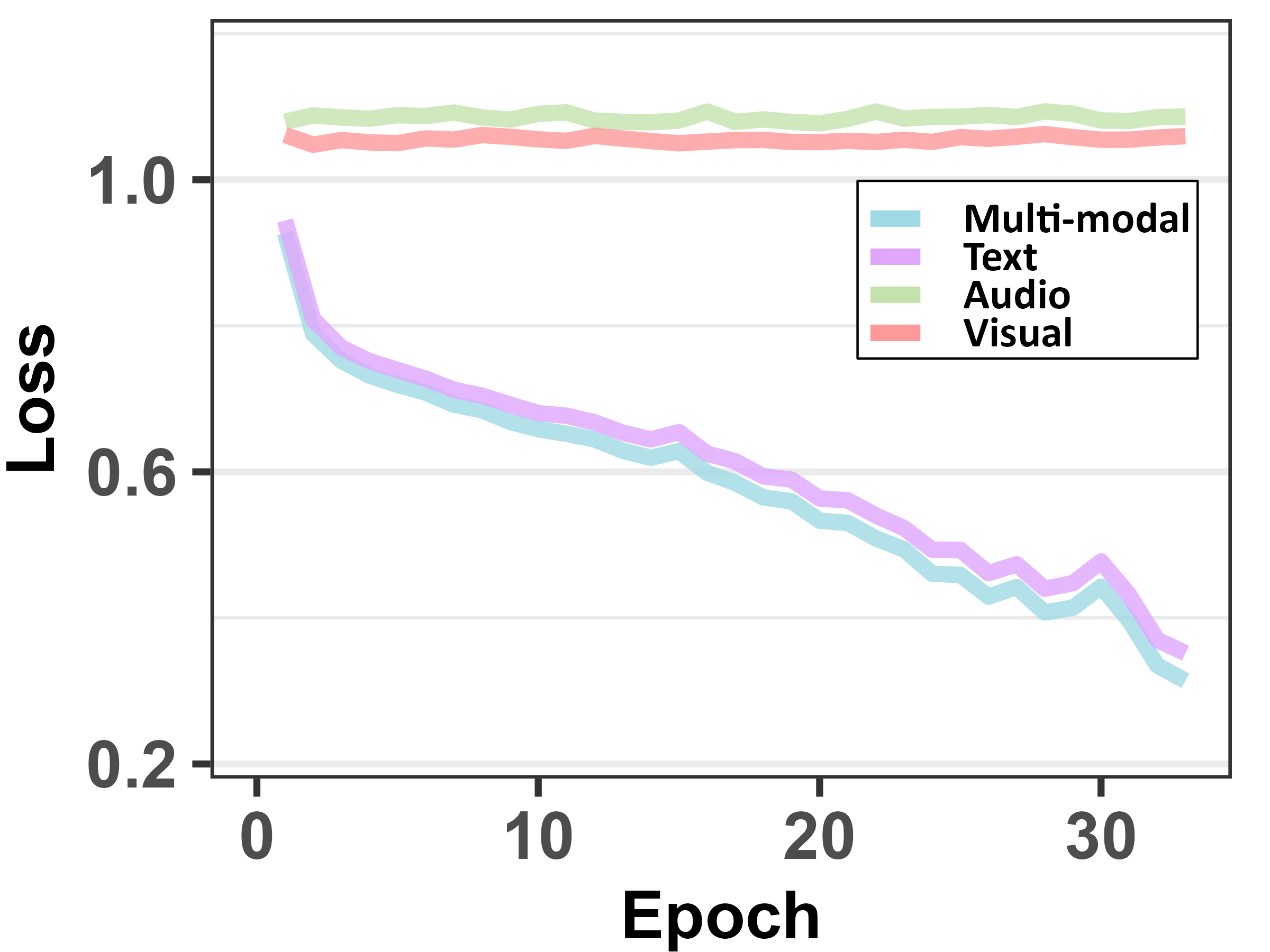}}
\caption{The phenomenon of modality competition is observed in the concatenation fusion method when applied to two datasets: CREMA-D with two modalities and MOSEI with three modalities. In CREMA-D, the learning process is primarily influenced by the audio modality, leading to insufficient learning of the visual modality. In MOSEI, the text modality takes control of multi-modal learning, causing challenges in updating the parameters of both the audio and visual modalities.}
\label{fig:process}
\end{figure}

\subsection{The Challenge of Multi-modal Learning} 

Current paradigms \textbf{\textit{synchronously}} optimize the objective across all modalities. In this setting, a joint gradient descent update will trigger the modality competition. To see this, we expand the update rule for each modality. Here, we denote the merged score function as:
\begin{equation}
    \Phi_M^{t}(x_i) = \sum_{k=1}^{M} W_k^t \cdot \mathcal{F}_k(\theta_k^t;m_i^k)
\end{equation}
Then the update for the $k$-th modality-specific parameters can be written as:
\begin{equation*}
    \small 
    \begin{split}
        \theta_k^{t+1} & = \theta_k^{t} - \eta\cdot\nabla_{\theta_k^{t}}\mathcal{L}( \Phi_M^{t}(x),y) \\
        & = \theta_k^{t} - \eta\cdot \frac{1}{N} \sum_{i=1}^{N} \left(\frac{\partial W_k \cdot \mathcal{F}_k(\theta_k^{t};m_i^k)}{\partial \theta_k^t}\right)^{\top} \\
        & {\color{white}= \theta_k^{t} - \eta\cdot \frac{1}{N} \sum_{i=1}^{N}\tilde{\mathcal{L}}\sum_{i=1}^{N}\tilde{\mathcal{L}}\sum_{i=1}^{n}\sum_{i=1}^{n}} \cdot\underbrace{\frac{\partial \ell\left(\Phi_{M}^{t}(x_i), y_i
        \right)}{\partial \Phi_M^{t}(x_i)}}_{\text{shared}}
    \end{split}
    \label{eq:update1}
\end{equation*}
\begin{equation*}
    \small
    \begin{split}
        W_k^{t+1} & = W_k^{t} - \eta\cdot\nabla_{W_k^{t}}\mathcal{L}( \Phi_M^{t}(x),y) \\
        & = W_k^{t} - \eta\cdot \frac{1}{N} \sum_{i=1}^{N} \left(\mathcal{F}_k(\theta_k^{t};m_i^k)\right)^{\top} \cdot \underbrace{\frac{\partial \ell(\Phi_M^{t}(x_i), y_i)}{\partial \Phi_M^{t}(x_i)}}_{\text{shared}},
    \end{split}
    \label{eq:update2}
\end{equation*}
where $\eta$ is the learning rate. The modality competition issue arises from the shared score gradient across modalities:
\begin{align}
    \frac{\partial \ell(\Phi_{M}^{t}(x_i), y_i)}{\partial \Phi_{M}^{t}(x_i)} 
    &= \sigma_i^t - y_i
\end{align}
where $\sigma_{i} \in \mathbb{R}^{Y}$ means the prediction score for the $i$-th example. Once the gradient for the shared score is small, the update for all modalities will be stuck simultaneously. 
A modality $k$ is said to have a consistent gradient with the shared scoring function  if $\frac{\partial \ell(\phi_k^{t}(x_i), y_i)}{\partial \phi_k^{t}(x_i)}$ strongly resembles $\frac{\partial \ell(\Phi_M^{t}(x_i), y_i)}{\partial \Phi_M^{t}(x_i)}$, where $\phi_k$ is the uni-modal score for modality $k$. If not, we consider modality $k$ to have an inconsistent gradient with the shared scoring function. If the modality $k$ has a consistent gradient with the shared score, then we can learn it well under this setting. On the opposite, if modality $k$ has an inconsistent gradient with the shared score, it will be stuck at bad local optimums, leading to performance degradation.  This phenomenon is depicted in Fig.\ref{fig:process}. To address this issue, our goal is to achieve reconcilement among modalities, where one can find a better trade-off between the exploitation of modality-specific patterns and the exploration of modality-invariance patterns.

Despite recent efforts to design various strategies ($S$ mentioned above) in multi-modal learning to avoid modality competition, only limited improvements can be achieved, given the nature of synchronous optimization. The limitations inspire us to explore a modality-alternating learning strategy.

\begin{figure*}
\centering  
\includegraphics[width=0.99\linewidth]{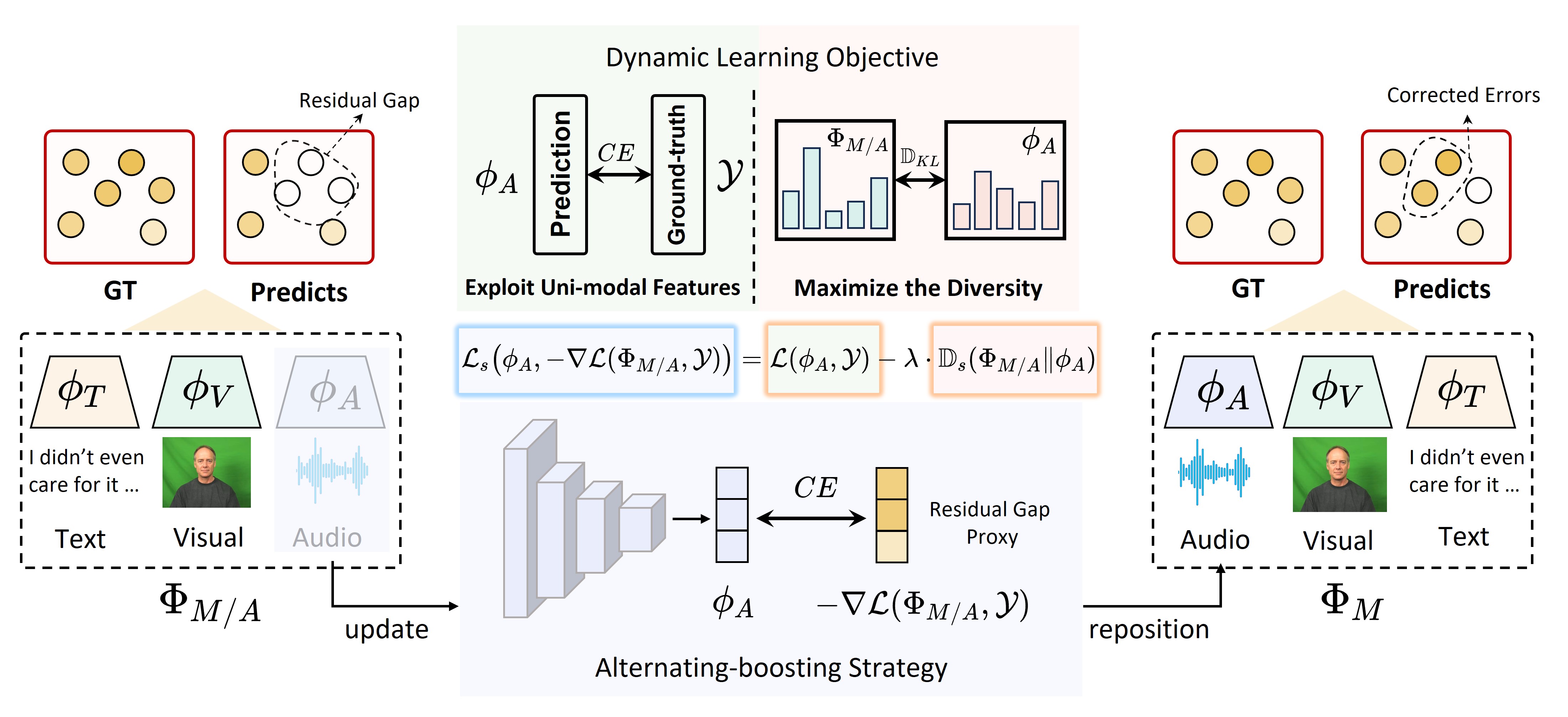}  
\caption{The overview of proposed ReconBoost. In round $s$, we pick up a specific modality learner to update and keep the others fixed. The updated modality learner can correct errors and enhance the overall performance.}  
\label{fig:pipline}  
\end{figure*}
\section{Methodology}
In this section, we propose a modality-alternating framework called ReconBoost.
The overall framework is illustrated in Fig.\ref{fig:pipline}. On top of the alternating update rule, we also incorporate a reconcilement regularization strategy to maximize the diversity between the current and historical models. 
Further details on ReconBoost will be discussed in the following.
\subsection{Modality-alternating Update with Dynamic Reconcilment}

\textbf{Notations.} Given $M$ modality-specific classifiers $\{W_k\}_{k=1}^{M}$, along with modality-specific feature extractors, $\{\phi_k(\vartheta_k)\}_{k=1}^{M}$ represents $M$ modality learners, where $\phi_k(\vartheta_k) = W_k\cdot \mathcal{F}_k(\theta_k)$. $\vartheta_k$ represents the parameters of the $k$-th modality learner and $\phi_k(\vartheta_k;m_i^k)\in \mathbb{R}^{Y}$. 

We first introduce the naive version of modality-alternating learning.

{\textbf{Step 1:}} Each time, we pick a specific modality learner $\phi_k$ to update, and keep the others fixed. The gradient rule is formalized as follows: 
\begin{align}
    \vartheta_k^{t+1} = \vartheta_k^{t} - \eta\cdot\nabla_{\vartheta_k^{t}}\mathcal{L}\left(\phi_k^{t}(m^{k}), y\right)
    \label{eq:as}
\end{align}
where 
\begin{equation}\small
\mathcal{L}\left(\phi_k^{t}(m^k), y\right) = \frac{1}{N}\sum_{i=1}^{N}\ell\left(\phi_k(\vartheta_k^t;m_i^k),y_i\right)   
\end{equation}
is the loss for the $k$-th modality. 

{\textbf{Step 2:}} After the alternating training procedure, multi-modal features are merged in the following way to produce the final score $\Phi_{M}(x_i) = \sum_{k=1}^{M}\phi_k(\vartheta_k;m_i^k)$.

When model updates are alternated, the gradients across different modalities are naturally disentangled from each other, alleviating the modality competition issue. While this approach ensures the exploitation of uni-modal features, it neglects the investigation of cross-modal diversity, limiting overall performance. It motivates us to design a more effective modality-specific supervised signal.

At each fixed time $s$ in the update in step 1), we explore reconcilement regularization  by introducing the following term in the loss:
\begin{equation*}\small
\mathbb{D}_{s}\left(\Phi_{M/k}(x_i),\phi_k (\vartheta_k;m_i^k)\right)
\end{equation*}
where $\Phi_{M/k}(x_i) = \sum_{j=1,j\neq k}^{M}\phi_j(\vartheta_j;m_i^j)$. Here, $\mathbb{D}_s$ could be regarded as a diversity measure between the current block being updated and the historical models in the updating sequence.
In pursuit of a dynamical reconcilement, in round $s$ in Step 1), we turn to use the following objective:
\begin{equation}
    \small
    \begin{split}  
       \tilde{\mathcal{L}}_s(\phi_k(m^k),y) & = \frac{1}{N}\sum_{i=1}^{N}\left [ \underbrace{\ell\left(\phi_k(\vartheta_k;m_i^k),y_i\right)}_{\text{agreement term}} \right. \\
        & \left. - \lambda\cdot \underbrace{\mathbb{D}_{s}\left(\Phi_{M/k}(x_i),\phi_k (\vartheta_k;m_i^k)\right)}_{\text{reconcilement regularization term}} \right ]
    \end{split}
    \label{eq:kl}
\end{equation}
In this new formulation, the loss function is no longer the same. In each round, we try to dynamically maintain the trade-off between \textbf{the agreement item} to align the overall predictor with the ground truth and \textbf{the reconcilement regularization term} to leverage complementary information between modalities. The parameter $\lambda$ is a trade-off coefficient. The exploration of the impact of $\lambda$ on the performance is presented in Sec.\ref{sec:abstudy} ablation experiments.

\subsection{Connection to the Boosting Strategy: Theoretical Guanratee}
At first glance, the dynamical variation of the loss function makes the optimization property of ReconBoost unclear.  To further explore its theoretical foundation, we investigate the connection with the well-known Gradient-Boosting (GB) method \cite{freund1995boosting,gb,freund1996experiments,freund1997decision}, which is a powerful boosting method for additive expansion of models. The theoretical result is shown as follows:

\begin{theorem} 
When the reconcilement regularization satisfies, 
\begin{equation*}\small
\begin{aligned}
    \lambda\cdot\nabla_{\phi_k}\mathbb{D}_{s}\left(\Phi_{M/k}(x_i),\phi_k (\vartheta_k;m_i^k)\right) = \nabla_{\phi_k}\ell\left(\phi_k (\vartheta_k;m_i^k),y_i\right) \\ 
      - \nabla_{\phi_k}\ell\left(\phi_k (\vartheta_k;m_i^k),-\nabla_{\Phi_{M/k}}\ell(\Phi_{M/k}(x_i),y_i)\right)  
\end{aligned}
\end{equation*}
It leads to equivalent optimization goals:
\begin{equation*}
\begin{aligned}
\nabla_{\vartheta_k}{\color{blue}\tilde{\mathcal{L}}}_s\left(\phi_k(m^k),{\color{blue} y}\right) \iff &\nabla_{\vartheta_k}{\color{blue}{\mathcal{L}}}{\left(\phi_{k}(m^k), \right.}\\
    &\quad { \left. {\color{blue}-\nabla_{\Phi_{M/k}} \ell(\Phi_{M/k}(x), y)}\right)}
\end{aligned}
\end{equation*}
\label{thm:main}
\end{theorem} 
\vskip -0.2in
\textbf{\textit{Please refer to App.\ref{app:proof} for the proof in detail.}}

Here, to better understand the generality of our method and theorem, we will consider the case where the optimization loss function is CE loss. As a corollary of the theorem we have:
\begin{corollary}
Let the reconcilement regularization be a KL divergence \cite{kl-divergence} function:
\begin{equation*}\small
\begin{aligned}
   \small{\mathbb{D}_{s}\left(\Phi_{M/k}(x_i),\phi_k (\vartheta_k;m_i^k)\right) = \mathbb{D}_{KL,s}\left(\Phi_{M/k}(x_i)\|\phi_k (\vartheta_k;m_i^k)\right)}
\end{aligned} 
\end{equation*}
Then,
\begin{equation*}
\begin{aligned}
\nabla_{\vartheta_k}{\tilde{\mathcal{L}}}_s\left(\phi_k(m^k),{y}\right) \iff &\nabla_{\vartheta_k}{{\mathcal{L}}}{\left(\phi_{k}(m^k), \right.}\\
    &\quad { \left. {-\nabla_{\Phi_{M/k}} \ell(\Phi_{M/k}(x), y)}\right)}
\end{aligned}    
\end{equation*}
where $\ell$ is the CE loss.
\label{cor:main}
\end{corollary}

Similar to the GB algorithm, optimizing the dynamic loss function $\tilde{\mathcal{L}}$ in ReconBoost consistently optimizes the original loss ${\mathcal{L}}$ with a progressively changing pseudo-label $-\nabla_{\Phi_{M/k}} \ell(\Phi_{M/k}(x), y)$. The pseudo-label is a gradient descent step at the space of $\Phi$ for the current time. The major difference from traditional GB is that we only employ the sum of the last updates for each modality, creating an \textbf{\textit{alternating-boosting strategy.}} This is a \textbf{selective} additive expansion of the gradient decent on the functional space. This could be considered an \textbf{implicit bias} when the weak learners in traditional GB are replaced with over-parametrized deep learning models. 

\subsection{Enhancement  Schemes}

In this subsection, we elaborate on two enhancement schemes in \name, memory consolidation regularization, and global rectification scheme.

\textbf{Memory Consolidation Regularization ($\mathsf{MCR}$).}
In contrast to GB, our alternating-boosting strategy preserves the newest learner for each modality while forgetting historical ones. Each updated modality learner fits the residual and effectively corrects errors from others. However, forgetting may result in modality-specific learners struggling with samples where others excel. To compensate for the discards, we propose $\mathsf{MCR}$ to enhance the performance of the modality-specific learner, formalized as:
\begin{equation}
\small
\begin{split}
&\mathcal{L}_{mcr}(- \nabla_{\phi_{k-1}} \ell(\phi_{k-1}(m^{k-1}), y), - \nabla_{\phi_{k}} \ell(\phi_{k}(m^k), y)) \\
&= \frac{1}{N}\sum_{i=1}^{N} \left\| \nabla_{\phi_{k}} \ell(\phi_{k}(m^k_i), y_i)-\nabla_{\phi_{k-1}} \ell(\phi_{k-1}(m^{k-1}_i), y_i) \right\| ^2
\end{split}
\label{eq:ga}
\end{equation}
where $\phi_{k-1}$ represents the previous modality learner. Intuitively, optimizing Eq.\ref{eq:ga} ensures that the predictions of $\phi_{k}$ will not be too far from that of $\phi_{k-1}$, avoiding excessive focus on errors and benefiting consolidating memory of the modality-specific learner.

\textbf{Global Rectification Scheme ($\mathsf{GRS}$).} Following the standard paradigm of boosting, only the parameters of the $k$-th weak learner get updated during step $k$ to greedily fit the residual, leaving the parameters of other learners unchanged. However, when dealing with modality learners implemented as over-parameterized neural networks in our alternating-boosting strategy, greedy learning strategy in the standard paradigm of boosting may cause the ensemble multi-modal learning model to fall in poor local minima easily, hindering the optimization of the objective. Drawing inspiration from \cite{grownet}, we introduce $\mathsf{GRS}$ to overcome the challenge. After each update of the modality learner, instead of keeping the parameters of the $k-1$ modality learners fixed, we allow their parameters to be updated:
\begin{equation}
    \vartheta_{m}^{t} = \vartheta_{m}^{t-1} - \eta\nabla_{\vartheta_{m}^{t-1}} \mathcal{L}(\Phi_{M}^{t-1}(x),y),\forall m\in[1,M]
\end{equation}
where $\Phi_M$ represents adding the updated $\phi_k$ to $\Phi_{M/k}$; $t$ means the $t$-th iteration in the rectification stage and $\eta$ is the learning rate.                                                           
In summary, these two schemes will enhance the performance of the proposed alternating-boosting strategy. Moreover, they provide insights into applying boosting techniques in the deep learning community.

\subsection{Final Goal}

Upon completing a cycle involving $M$ stages, we reach the overall optimization objective for our proposed method in Eq.\ref{eq:overall}.
\vskip -0.1in
\begin{equation}
    \small
    \begin{aligned}
        \mathcal{L}_{all} &= \underbrace{\sum_{k\in[1,M]}\mathcal{L}(\phi_k(m^k),y)}_{\text{agreement term}}
        - \underbrace{\lambda\sum_{k\in[1,M]}\mathbb{D}_{KL}(\Phi_{M/k}(x)\| \phi_k(m^k))}_{\text{KL-based reconcilement regularization term}} \\
        &+ \alpha \underbrace{\sum_{k\in[1,M]} \mathcal{L}_{mcr}(-\nabla_{\phi_{k-1}} \ell(\phi_{k-1}(m^{k-1}), y), -\nabla_{\phi_{k}} \ell(\phi_{k}(m^k), y))}_{\text{ $\mathsf{MCR}$ term}} \\
        &+ \underbrace{\sum_{k\in[1,M]}\mathcal{L}(\Phi_{M}(x),y)}_{\text{$\mathsf{GRS}$ term}}
    \end{aligned}
    \label{eq:overall}
\end{equation}

The pseudo-code of training ReconBoost is detailed in Alg.\ref{alg:algorithm1}. In lines $4$ to $7$, we calculate the dynamic modality-specific loss including the agreement term, KL-based reconcilement regularization term, and $\mathsf{MCR}$ term to update the $k$-th modality learner. After updating, in lines $10$ to $13$, we employ the $\mathsf{GRS}$ to perform global rectification. The process then continues with the update of the next modality learner.

\begin{algorithm}[tb]
\caption{ReconBoost Algorithm}
\label{alg:algorithm1}
\LinesNumbered
\KwIn{Observations $\mathcal{D}_{train}$, iterations of each stage $T$, lr in alternating-boosting stage $\gamma$, lr in global rectification stage $\eta$}
\KwOut{Well-trained model $\Phi_{M}$}
\Repeat{\text{converge}}{
    $\textcolor{orange}{\rhd}$ \textcolor{orange}{Alternating-boosting Strategy}

    In round $s$, pick up a modality-specific learner $\phi_k,k\in[1,M]$ to be updated in order\;
    
    \For{$t=0$ {\bfseries to} $T-1$}{
        Sample $\forall \{x_i,y_i\} \in \mathcal{D}_{train}$\;
        
        \text{Calculate modality-specific loss} $\ell_{A}\left( \phi_k^{t}\right) =\ell(\phi_k^{t}(m_i^k),y_i) -\lambda \mathbb{D}_{KL,s}+ \alpha \ell_{mcr}$\;
    
        \text{Update} $\vartheta_k^{t+1} = \vartheta_k^{t} - \gamma \cdot \nabla_{\vartheta_k^{t}}\ell_{A}\left( \phi_k^{t} \right)$\;
    }
    
    Add the model $\phi_k^{T}$ into the $\Phi_{M/k}$, denoted as $\Phi_{M,s}$\;
    
    $\textcolor{orange}{\rhd}$ \textcolor{orange}{Global Rectification Scheme} 
    
    \For{$t=0$ {\bfseries to} $T-1$}{
        Sample $\forall \{x_i,y_i\} \in \mathcal{D}_{train}$\;
    
        Calculate loss: $\ell_{G}(\Phi_{M,s}^{t}) =\ell(\sum_{k=1}^{M}\phi_k^{t}(m_i^k),y_i)$\;
    
        Update all modality learners $\forall m\in[1,M]$ $\vartheta_m^{t+1} = \vartheta_m^t - \eta\cdot\nabla_{\vartheta_m^t}\ell_{G}(\Phi_{M,s}^{t})$\;
    }
}
\Return $\Phi_{M}$.
\end{algorithm}
\section{Experiments}
In this section, we provide the empirical evaluation across a wide range of multi-modal datasets to show the superior performance of ReconBoost. \textbf{\textit{Due to space limitations, please refer to App.\ref{app:exp_setting} and \ref{app:exp_analysis} for an extended version.}}

\subsection{Experimental Setup}

\textbf{Dataset Descriptions.} We conduct empirical experiments on several common multi-modal benchmarks. Specifically,  \textbf{AVE} \cite{ave} dataset is designed for audio-visual event localization and includes $28$ event classes. \textbf{CREMA-D} \cite{cremad} is an audio-visual video dataset for speech emotion recognition including 6 emotion classes. \textbf{ModelNet40} \cite{mn40} a large-scale $3$-D model dataset, with the front and rear view to classify the object, following \cite{greedy} and \cite{UMT}. \textbf{MOSEI} \cite{mosei}, \textbf{MOSI} \cite{mosi}, and \textbf{CH-SIMS} \cite{ch-sims} are multi-modal sentiment analysis datasets including three modalities, namely audio, image, and text. We defer the detailed introductions of these datasets to App.\ref{app:dataset}

\textbf{Competitors.} To demonstrate the effectiveness of the proposed method, we compare it with some recent multi-modal learning methods that focus on alleviating modality competition. These competitors include \textbf{G-Blending} \cite{g-blending}, \textbf{OGM-GE} \cite{OGM_GE_2022_CVPR}, \textbf{PMR} \cite{PMR_CVPR_23}, \textbf{UME} \cite{UMT} and \textbf{UMT} \cite{UMT}. We also include the \textbf{Concat}enation fusion method and \textbf{Uni}-modal methods. Detailed explanations of these competitors will be provided in App.\ref{app:comp}.

\textbf{Implementation Details.} All experiments are conducted on GeForce RTX 3090 and all models are implemented with Pytorch \cite{paszke2017automatic}. Specifically, for the AVE, CREMA-D and ModelNet40 datasets, we adopt ResNet-18 \cite{resnet} as the backbone and modify the input channel according to the size of different modalities. For MOSEI, MOSI, and SIMS datasets, we conduct experiments with fully customized multimodal features extracted by the MMSA-FET toolkit. The uni-modal models are similar to \cite{emotion-dnn}. We adopt SGD \cite{gradient_descent} as the optimizer. Specific data preprocessing, network design and optimization strategies are provided in App.\ref{sec:implement_details}.

\begin{table}[]
\renewcommand\arraystretch{1.0}
\center
\caption{Performance comparisons on AVE, CREMA-D, and ModelNet40 in terms of Acc(\%). In the MN40 dataset, following UMT \cite{UMT}, we use different views, so there are no prediction results of uni-audio modality, denoted as '-'.}
\vskip 0.1in
\begin{center}
    \begin{tabular}{c|ccc}
    \toprule
    \multicolumn{1}{c}{\textbf{Method}} & \textbf{AVE} & \textbf{CREMA-D} & \textbf{MN40} \\
    \midrule
    AudioNet & 59.37 & 56.67 & - \\
    VisualNet & 30.46 & 50.14 & 80.51 \\
    Concat Fusion & 62.68 & 59.50  & 83.18 \\
    G-Blending  & 62.75 & \cellcolor[rgb]{ .996,  .957,  .933}63.81 & \cellcolor[rgb]{ .996,  .969,  .949}84.56 \\
    OGM-GE & \cellcolor[rgb]{ 1,  .996,  .992}62.93 & \cellcolor[rgb]{ .992,  .941,  .906}65.59 & \cellcolor[rgb]{ .996,  .945,  .91}85.61 \\
    PMR   & \cellcolor[rgb]{ .996,  .965,  .945}64.20 & \cellcolor[rgb]{ .992,  .937,  .898}66.10 & \cellcolor[rgb]{ .992,  .929,  .89}86.20 \\
    UME   & \cellcolor[rgb]{ .988,  .902,  .843}66.92 & \cellcolor[rgb]{ .988,  .914,  .863}68.41 & \cellcolor[rgb]{ .996,  .949,  .922}85.37 \\
    UMT   & \cellcolor[rgb]{ .984,  .882,  .816}67.71 & \cellcolor[rgb]{ .988,  .886,  .82}70.97 & \cellcolor[rgb]{ .98,  .839,  .745}90.07 \\
    \midrule
    \textbf{Ours} & \cellcolor[rgb]{ .973,  .796,  .678}\textbf{71.35} & \cellcolor[rgb]{ .973,  .796,  .678}\textbf{79.82} & \cellcolor[rgb]{ .973,  .796,  .678}\textbf{91.78} \\
    \bottomrule
    \end{tabular}%
\end{center}
\vskip -0.1in
\label{tab:CREMA-D_ave}
\end{table}


\subsection{Overall Performance}

The experimental results are presented in Tab.\ref{tab:CREMA-D_ave} and Tab.\ref{tab:three}. Our proposed methods consistently outperform all competitors significantly across all datasets, underscoring the efficacy of our approach. In Tab.\ref{tab:CREMA-D_ave}, within a dataset featuring two modalities, all multi-modal learning methods exhibit improvements compared to the naive concatenation fusion method. This observation confirms the existence of modality competition in multi-modal joint learning, demonstrating the effective alleviation of modality competition by the compared methods. Specifically, given a well-trained AudioNet, our method achieves the most significant improvements when incorporating the visual modality, which justifies that our method can effectively make the most of cross-modal information.

In contrast to prior studies, we also evaluate the effectiveness of various modulation strategies on tri-modality datasets. Some earlier strategies \cite{OGM_GE_2022_CVPR,PMR_CVPR_23} focused on mitigating competition between two modalities and lacked generalization to multiple modalities. For these methods, we test combinations of different modalities and select better models for presentation. Our approach treats a multi-modal learning framework as a generalized ensemble model and demonstrates robust generalization across multiple modalities.

To further demonstrate ReconBoost's adaptability in broader contexts, we applied it to the retrieval task, a crucial area within computer vision. We assessed its performance using the Mean Average Precision (MAP) metric on the CREMA-D as shown in Tab.\ref{tab:main_retrieval}. The detailed comparison results are provided in App.\ref{sec:app_retrieval}

\begin{table}[]
\renewcommand\arraystretch{1.0}
\center
\caption{Performance comparisons on MOSEI, MOSI, and CH-SIMS datasets in terms of Acc(\%).}
\vskip 0.1in
\begin{center}
\begin{tabular}{c|ccc}
    \toprule
    \multicolumn{1}{c}{\textbf{Method}} & \textbf{MOSEI} & \textbf{MOSI} & \textbf{CH-SIMS} \\
    \midrule
    AudioNet & 52.29 & 54.81 & 58.20 \\
    VisualNet & 50.35 & 57.87 & 63.02 \\
    TextNet & 66.41 & \cellcolor[rgb]{ .984,  .992,  .996}75.94 & 70.45 \\
    Concat Fusion & \cellcolor[rgb]{ .969,  .98,  .992}66.71 & \cellcolor[rgb]{ .949,  .973,  .988}76.23 & \cellcolor[rgb]{ .918,  .953,  .98}71.55 \\
    G-Blending  & \cellcolor[rgb]{ .941,  .965,  .984}66.93 & \cellcolor[rgb]{ .925,  .953,  .98}76.45 & \cellcolor[rgb]{ .918,  .953,  .98}71.55 \\
    OGM-GE & \cellcolor[rgb]{ .973,  .984,  .992}66.67 & \cellcolor[rgb]{ .976,  .988,  .996}76.01 & \cellcolor[rgb]{ .953,  .973,  .988}71.10 \\
    PMR   & 66.41 & \cellcolor[rgb]{ .965,  .98,  .992}76.12 & \cellcolor[rgb]{ .969,  .98,  .992}70.90 \\
    UME   & 63.88 & \cellcolor[rgb]{ .863,  .918,  .965}76.97 & \cellcolor[rgb]{ .902,  .941,  .976}71.77 \\
    UMT   & \cellcolor[rgb]{ .929,  .957,  .984}67.04 & 75.80  & \cellcolor[rgb]{ .918,  .953,  .98}71.55 \\
    \midrule
    \textbf{Ours} & \cellcolor[rgb]{ .741,  .843,  .933}\textbf{68.61} & \cellcolor[rgb]{ .741,  .843,  .933}\textbf{77.96} & \cellcolor[rgb]{ .741,  .843,  .933}\textbf{73.88} \\
    \bottomrule
    \end{tabular}%
\end{center}
\vskip -0.1in
\label{tab:three}
\end{table}

\begin{table}[]
  \centering
  \caption{Performance comparisons on the AVE and CREMA-D datasets in terms of mAP(\%).}
  \vskip 0.1in
    \begin{tabular}{c|ccc}
    \toprule
    \multicolumn{1}{c}{\textbf{Method}} & Overall & Audio & Visual \\
    \midrule
    Concat Fusion & 36.43  & 34.71  & 20.08  \\
    OGM-GE & \cellcolor[rgb]{ 1,  .996,  .984}38.50  & \cellcolor[rgb]{ 1,  .984,  .937}36.59  & \cellcolor[rgb]{ 1,  .996,  .976}24.42  \\
    PMR   & \cellcolor[rgb]{ 1,  .996,  .976}39.34  & \cellcolor[rgb]{ 1,  .98,  .925}36.97  & \cellcolor[rgb]{ 1,  .996,  .973}25.10  \\
    UME   & \cellcolor[rgb]{ 1,  .996,  .973}40.02  & \cellcolor[rgb]{ 1,  .98,  .922}37.12  & \cellcolor[rgb]{ 1,  .988,  .941}30.45  \\
    UMT   & \cellcolor[rgb]{ 1,  .988,  .949}42.58  & \cellcolor[rgb]{ 1,  .992,  .969}35.65  & \cellcolor[rgb]{ 1,  .984,  .929}32.41  \\\midrule
    \textbf{Ours} & \cellcolor[rgb]{ 1,  .949,  .8}\textbf{60.52 } & \cellcolor[rgb]{ 1,  .949,  .8}\textbf{40.58 } & \cellcolor[rgb]{ 1,  .949,  .8}\textbf{54.26 } \\
    \bottomrule
    \end{tabular}%
  \label{tab:main_retrieval}%
\end{table}%

\textbf{Applicable to Other Fusion Schemes.} 
Our ReconBoost framework can be easily combined with several decision-level fusion methods, such as QMF \cite{dynamic_fusion} and TMC \cite{TMC}. Additionally, we benchmark against two straightforward baselines, Learnable Weighting (LW) and Naive Averaging (NA). To ensure fairness, we also included complex feature-based fusion, specifically MMTM \cite{mmtm}, into our main competitors: OGM-GE, PMR, and UMT. As shwon in Tab.\ref{tab:main_fusion}, our method consistently outperforms others, highlighting the potential of more flexible fusion strategies to enhance performance. This reaffirms the effectiveness of ReconBoost. The detailed comparison results and analysis are provided in App.\ref{sec:app_fusion}.

\begin{table}[]
  \centering
  \caption{Performance comparisons on the AVE and CREMA-D dataset in terms of Acc(\%) with different fusion strategies. $\dagger$ indicates that MMTM is applied.}
  \vskip 0.1in
    \begin{tabular}{c|cc}
    \toprule
    \multicolumn{1}{c}{\textbf{Method}} & \textbf{AVE} & \textbf{CREMA-D} \\
    \midrule
    OGM\_GE $\dagger$  & 66.14  & 69.83  \\
    PMR $\dagger$ & 67.72  & 70.14  \\
    UMT $\dagger$ & 70.16  & 74.35  \\ \midrule
    \textbf{Ours} + \textit{NA} & \cellcolor[rgb]{ .902,  .941,  .976}71.35  & 79.82  \\
    \textbf{Ours} + \textit{LW} & \cellcolor[rgb]{ .812,  .886,  .953}72.40  & \cellcolor[rgb]{ .945,  .969,  .988}80.11  \\
    \textbf{Ours} + \textit{TMC} & \cellcolor[rgb]{ .765,  .859,  .941}72.96  & \cellcolor[rgb]{ .827,  .898,  .957}80.68  \\
    \textbf{Ours} + \textit{QMF} & \cellcolor[rgb]{ .741,  .843,  .933}\textbf{73.20} & \cellcolor[rgb]{ .741,  .843,  .933}\textbf{81.11} \\
    \bottomrule
    \end{tabular}%
  \label{tab:main_fusion}%
\end{table}%

\begin{table}[]
\renewcommand\arraystretch{1.1}
\center
\caption{Performance of the encoders trained by Uni-modal, Concatenation fusion,  OGM-GE,  UMT, and Ours in terms of Acc(\%). We evaluate the encoders of all methods by training linear classifiers on them.}
\vskip 0.1in
\begin{center}
\begin{small}
    \begin{tabular}{ccccc}
    \toprule
    \multirow{2}[4]{*}{\textbf{Method}} & \multicolumn{2}{c}{\textbf{CREMA-D}} & \multicolumn{2}{c}{\textbf{AVE}} \\
\cmidrule{2-5}          & Visual & Audio & Visual & Audio \\
    \midrule
    Uni-train & \cellcolor[rgb]{ .89,  .941,  .855}50.14 & \cellcolor[rgb]{ .925,  .961,  .902}56.67 & \cellcolor[rgb]{ .906,  .949,  .875}30.46 & \cellcolor[rgb]{ .851,  .918,  .8}59.37 \\
    Concat Fusion & 26.81 & 54.86 & 23.96 & 55.47 \\
    OGM-GE & \cellcolor[rgb]{ .992,  .996,  .988}29.17 & \cellcolor[rgb]{ .98,  .988,  .973}55.42 & \cellcolor[rgb]{ .98,  .988,  .973}25.52 & \cellcolor[rgb]{ .961,  .98,  .949}56.51 \\
    PMR   & \cellcolor[rgb]{ .992,  .996,  .988}29.21 & \cellcolor[rgb]{ .973,  .984,  .961}55.60 & \cellcolor[rgb]{ .969,  .984,  .957}26.30 & \cellcolor[rgb]{ .933,  .965,  .914}57.20 \\
    UMT   & \cellcolor[rgb]{ .91,  .953,  .882}45.69 & \cellcolor[rgb]{ .851,  .922,  .804}58.47 & \cellcolor[rgb]{ .894,  .945,  .859}31.25 & \cellcolor[rgb]{ .796,  .89,  .733}60.70 \\
    \midrule
    \textbf{Ours} & \cellcolor[rgb]{ .776,  .878,  .706}\textbf{73.01} & \cellcolor[rgb]{ .776,  .878,  .706}\textbf{60.23} & \cellcolor[rgb]{ .776,  .878,  .706}\textbf{39.06} & \cellcolor[rgb]{ .776,  .878,  .706}\textbf{61.20} \\
    \bottomrule
    \end{tabular}%
\end{small}
\end{center}
\vskip -0.1in
\label{tab:uni}
\end{table}

\subsection{Quantitative Analysis}

\textbf{Modality-specific Encoder Evaluation.} We evaluate the encoders of Concatenation fusion, OGM-GE, PMR, UMT, and Ours by training linear classifiers on top of them. As shown in Tab.\ref{tab:uni}, in most methods, the dominant modality (audio modality) encoder can achieve comparable performance compared with its uni-modal counterpart, however, the weak modality (visual modality) encoder is far behind. Uni-modal information remains underutilized, and uni-modal features suffer corruption during joint training. For UMT, employing a uni-modal distillation strategy aids in exploiting sufficient uni-modal features, enabling some encoders of UMT to slightly outperform their uni-modal counterparts. However, distilled knowledge will be slightly corrupted in the fusion due to modality competition. 

Compared to them, the encoders trained by our proposed method achieve significant improvements. Benefiting from the alternating-learning paradigm, ReconBoost can avoid modality competition and ensure sufficient exploitations of uni-modal features. Furthermore, the innovative reconcilement regularization term  effectively leverages complementary information between modalities. Our method's encoders achieve  remarkable performance, which surpasses that of the uni-trained model.

\begin{figure}[]  
\centering  
\includegraphics[width=0.99\linewidth]{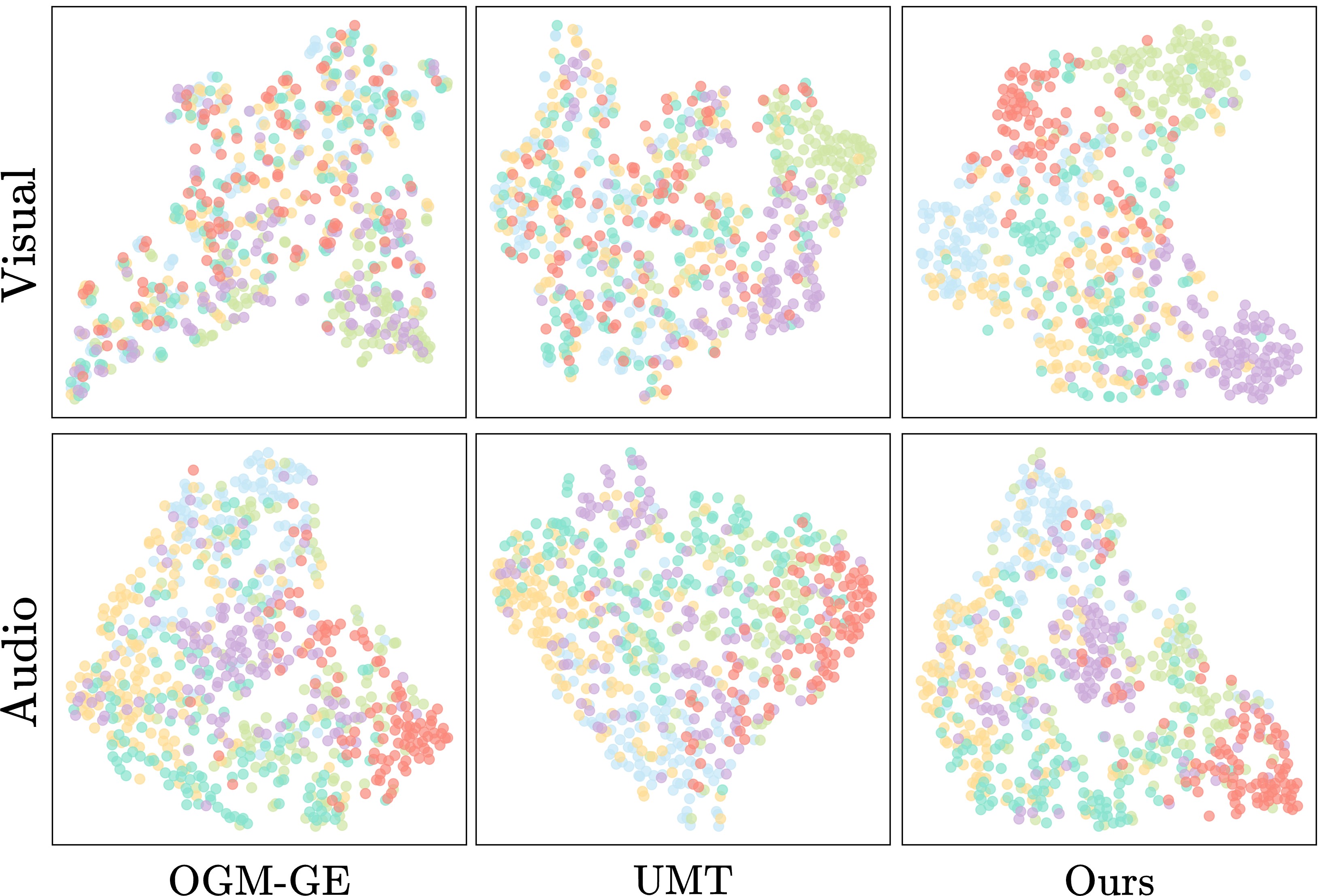}
\caption{The visualization of the modality-specific feature among different competitors in the CREMA-D dataset by using the t-SNE method \cite{tsne}.}  
\label{fig:tsne} 
\end{figure}

Fig.\ref{fig:tsne} shows the 2D embeddings of modality-specific features. In other methods, modality competition still exists. The features of the visual modality scatter randomly, reflecting low feature quality. Our approach focuses on improving the quality of latent features for each modality. Distinct clusters within each modality further highlight its effectiveness in reducing modality competition compared to other methods. The detailed comparison results are provided in App.\ref{app:tsne}.

\begin{figure}[]  
\centering  
\subfigure[MIR]{  
\includegraphics[width=0.46\linewidth]{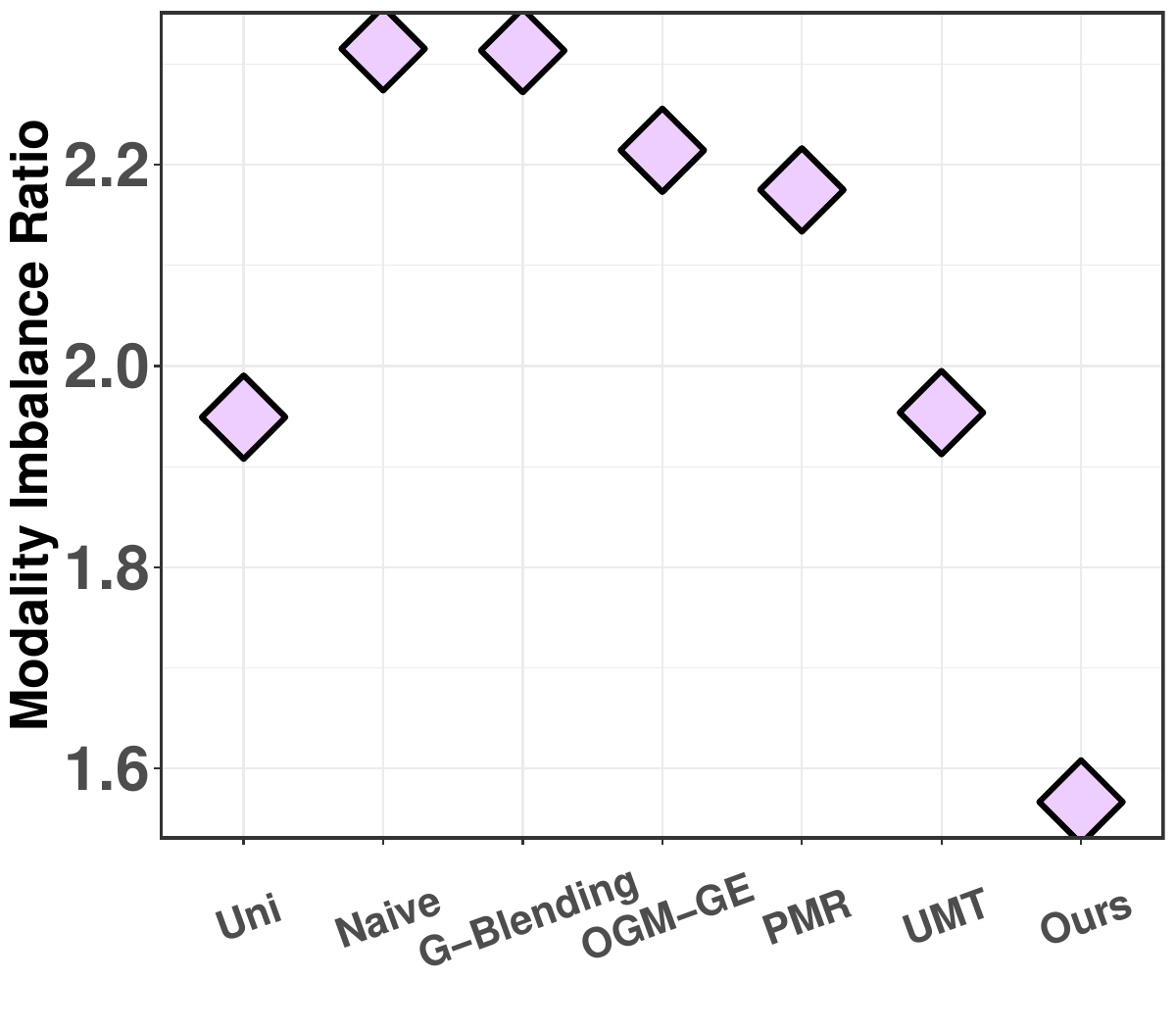}  
\label{fig:mir} 
}  
\subfigure[DMC]{  
\includegraphics[width=0.46\linewidth]{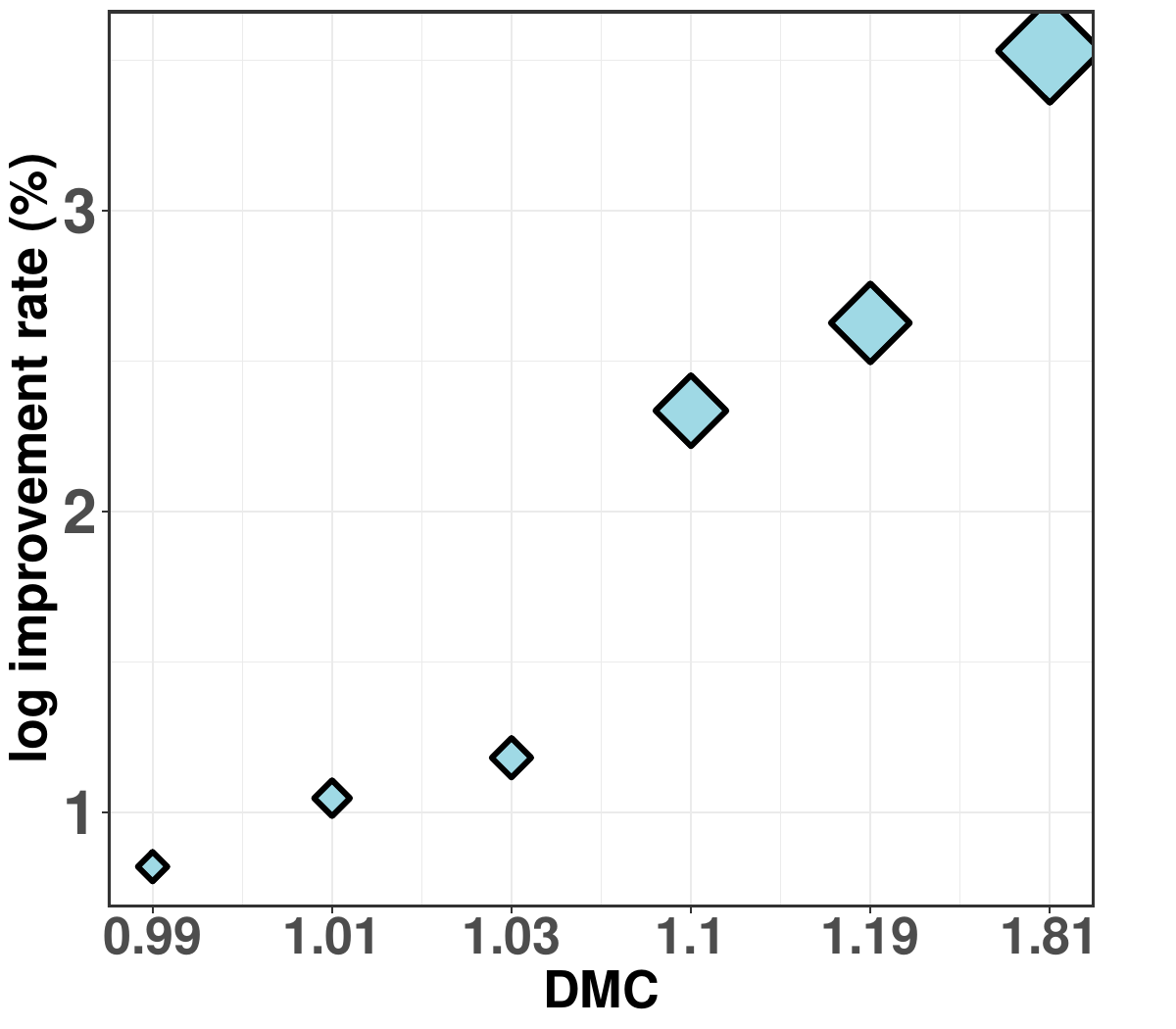}  
\label{fig:dmc}
}  
\caption{Quantitative analysis of modality competition. (a) Modality imbalance ratio (MIR) for all competitors on the AVE dataset. (b) The correlation between the DMC in the concatenation fusion method and the improvement of our method is consistent across all datasets.}  
\end{figure}

\textbf{Modality Competition Analysis.} Modality competition worsens the performance gap between modalities. To quantify this competition, we first measure the performance gap between modalities using the \textbf{modality imbalance ratio} (MIR). Moving further, we define the MIR of multi-modal learning methods relative to that of uni-modal learning as the degree of modality competition  (DMC). Fig. \ref{fig:mir} summarizes the modality imbalance ratio for all competitors on the AVE dataset. Although the MIR of various competitors is lower than that of naive concatenation fusion, it remains higher than the MIR under uni-modal learning, indicating the persistent challenge of modality competition. In contrast, our method effectively avoids modality competition, as revealed by the results. Fig.\ref{fig:dmc} illustrates the DMC value of the concatenation fusion method across all datasets. Notably, as the degree of modality competition rises, so does the improvement our method offers. The in-depth analysis regarding the phenomena are shown in App.\ref{app:modality_competition}.

\begin{figure}[]  
\centering  
\subfigure[Audio on CREMAD]{  
\includegraphics[width=0.46\linewidth]{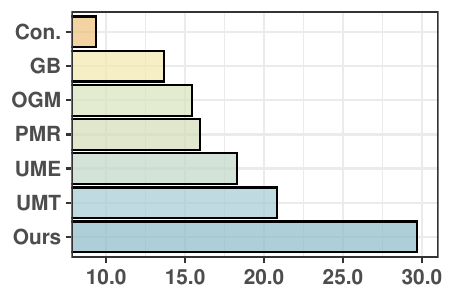}  
}  
\subfigure[Front View on MN40]{  
\includegraphics[width=0.46\linewidth]{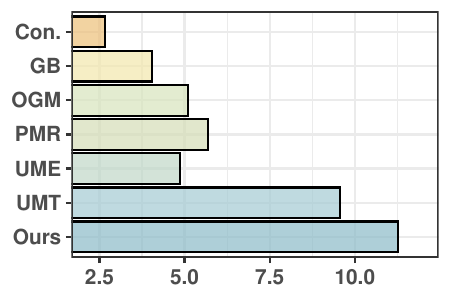}  
}  
\caption{Quantitative analysis of task relevant mutual information in audio modality on the CREMA-D dataset and in front view on the MN40 dataset.}  
\label{fig:Mutual.Info.}
\end{figure}

\textbf{Mutual Information Analysis.} We quantify task-relevant mutual information \cite{factor-contrastive-learning} in different multi-modal models. Firstly, we decompose the mutual information into shared information and modality-specific unique information. When only one modality $\mathcal{X}_1$ exists, the uni-modal only contains unique information. Then, in a multi-modal setting, maximize the information that $\mathcal{X}_2$ can bring becomes the key to improving the performance of multi-modal learning algorithms. We measure the information that $\mathcal{X}_2$ can bring using the difference in accuracy between using the multi-modal approach and the uni-modal model. As shown in Fig.\ref{fig:Mutual.Info.}, we evaluate it among different competitors on two benchmarks and our method consistently outperforms others, demonstrating the potential of maximizing the useful information in each modality. The in-depth analysis are provided in App.\ref{sec:app_MI}.

\subsection{Ablation Study}
\label{sec:abstudy}

\begin{figure}[]
\centering  
\subfigure[CREMA-D]{
\label{fig:ave}
\includegraphics[width=0.47\columnwidth]{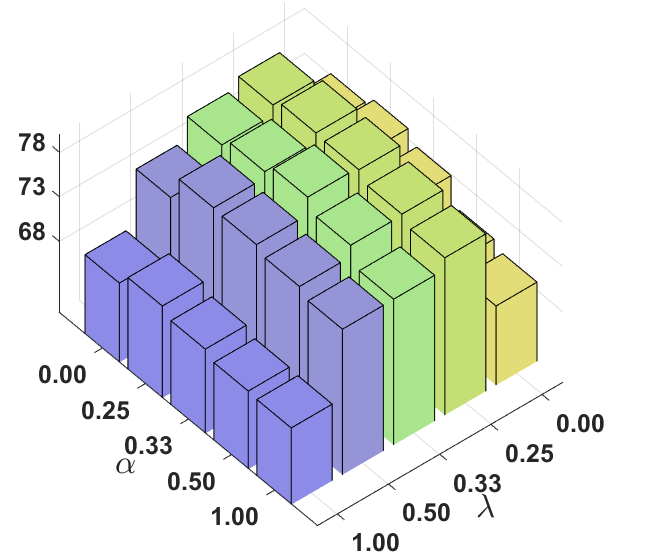}}
\subfigure[MOSEI]{
\label{fig:CREMA-D}
\includegraphics[width=0.47\columnwidth]{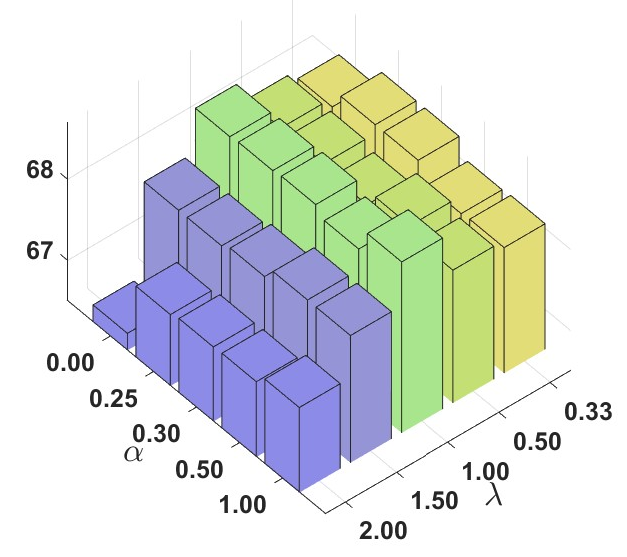}}
\caption{Sensitivity analysis about $\lambda$ and $\alpha$ on CREMA-D and MOSEI datasets.}
\label{fig:lambda_alpha}
\end{figure}

\textbf{Sensitivity analysis of $\lambda$.} Fig.\ref{fig:lambda_alpha} demonstrates the performance of ReconBoost with varying $\lambda$ on CREMA-D and MOSEI. We observe that a proper $\lambda$ could extract complementary information and significantly improve the performance. Leveraging $\lambda$ too aggressively may hurt the performance since excessive disagreement with others will damage modality-specific prediction accuracy. The detailed comparison results are provided in App.\ref{app:lambda}.

\textbf{Impact of Memory Consolidation Scheme.} Fig.\ref{fig:lambda_alpha} also explores the role of the $\alpha$ parameter in demonstrating the efficacy of the $\mathsf{MCS}$. Keeping $\lambda$ constant, we note that adjustments in $\alpha$ yield marginal yet meaningful improvements in performance. Given that $\lambda$ primarily governs the level of agreement, its adjustment can significantly enhance memory consolidation in modality-specific learning. This suggests that while $\lambda$ offers a broader range of manipulation for performance enhancement, fine-tuning with $\alpha$ allows for more precise and subtle improvements.

\begin{figure}[]
\centering
\subfigure[AVE Dataset]{  
\includegraphics[width=0.45\linewidth]{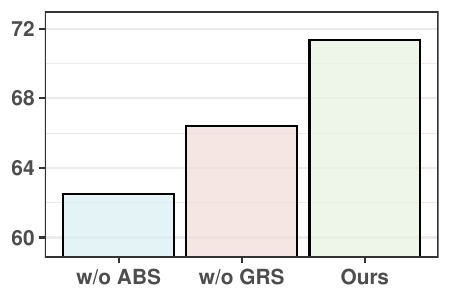}  
}  
\subfigure[CREMA-D Dataset]{  
\includegraphics[width=0.45\linewidth]{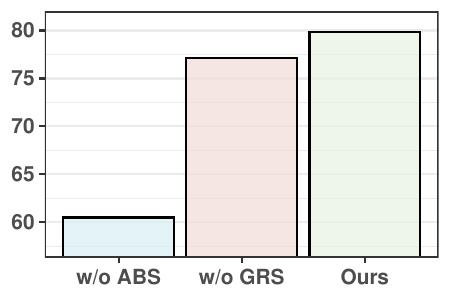}  
}  
\caption{Abation study of global rectification scheme ($\mathsf{GRS}$) on CREMA-D and AVE Dataset.}  
\label{fig:urs}
\end{figure}

\textbf{Effect of Global Rectification Scheme.} 
Fig.\ref{fig:urs} illustrates the effectiveness of the global rectification scheme by comparing w/o $\mathsf{GRS}$ and Ours. $\mathsf{GRS}$ facilitates the optimization of the multi-modal learning objective, preventing the ensemble model from falling into unfavorable local minima. Even without $\mathsf{GRS}$, our model achieves relatively good results, demonstrating that our alternating-boosting strategy effectively promotes the optimization of the objective.

\subsection{Convergence}

\begin{figure}[]
\centering  
\subfigure[AVE Dataset]{
\label{fig:con-ave}
\includegraphics[width=0.47\columnwidth]{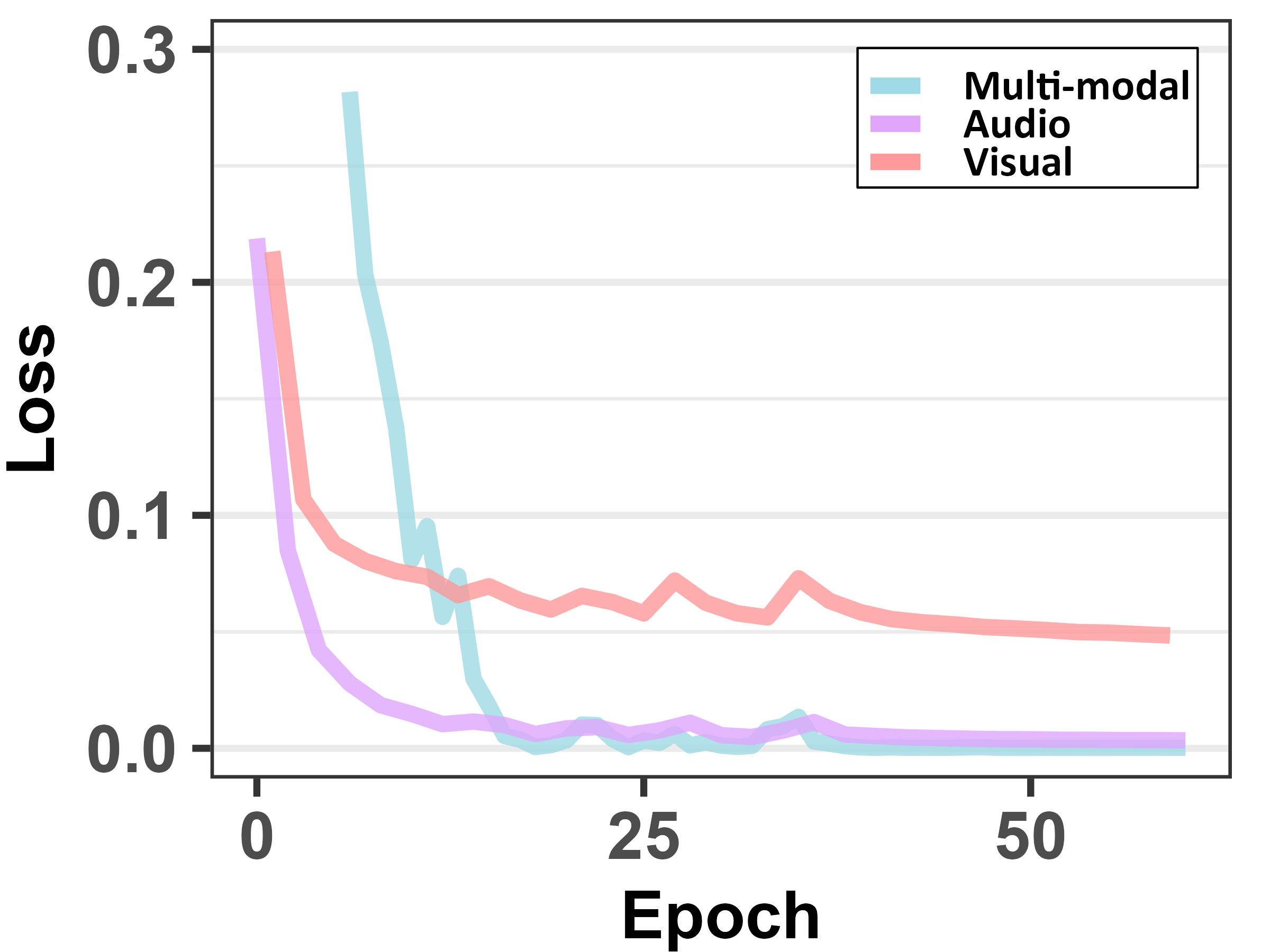}}
\subfigure[CREMA-D Dataset]{
\label{fig:con-CREMA-D}
\includegraphics[width=0.47\columnwidth]{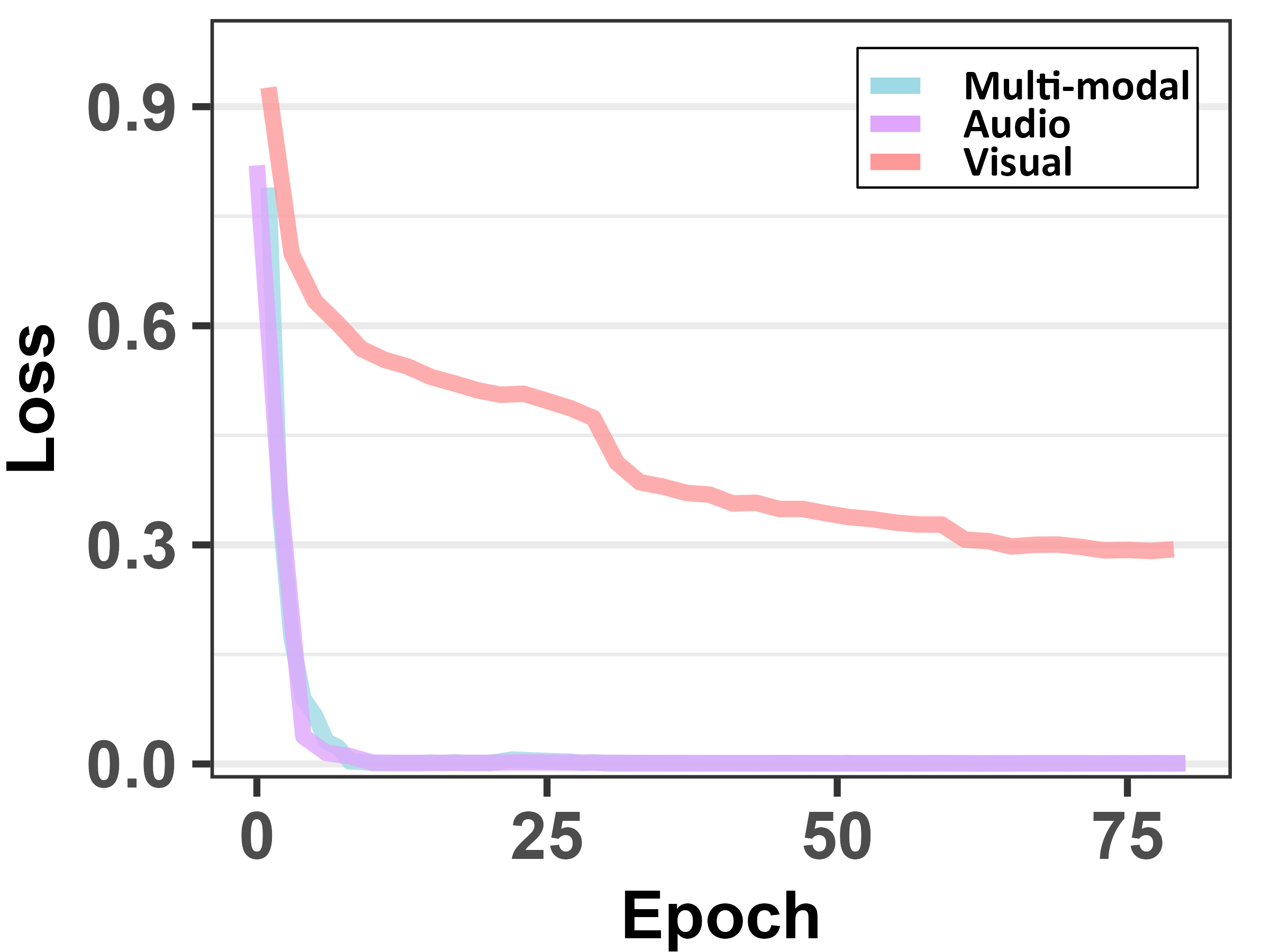}}
\caption{Convergence results of ReconBoost on AVE and CREMA-D Dataset.}
\label{fig:converge}
\end{figure}

We present the convergence results on two benchmark datasets during the training process, including AVE and  CREMA-D datasets. The performance results are shown in Fig.\ref{fig:converge}. For ReconBoost, the updates of all modality learners are alternated, and the gradients across different modalities are naturally disentangled from each other. Therefore, the modality-specific loss curve descends without getting stuck.

\vskip -0.1in
\section{Conclusion}

In this paper, we propose an effective multi-modal learning method based on an alternating learning paradigm to address the modality competition problem. Our method achieves a reconciliation between the exploitation of uni-modal features and the exploration of cross-modal interactions, with the crucial idea of incorporating a KL divergence based reconcilement regularization term. We have proven that optimizing modality-specific learners with this regularization is equivalent to the classic gradient-boosting algorithm. Therefore, the updated modality learner can fit the residual gap and promote the overall performance. We discard historical learners and only preserve the newest learners, forming an alternating-boosting strategy. Finally, the experiment results over a range of multi-modal benchmark datasets showcase significant performance improvements, affirming the effectiveness of the proposed method.

\section*{Acknowledgements}
This work was supported in part by the National Key R\&D Program of China under Grant 2018AAA0102000, in part by the National Natural Science Foundation of China: 62236008, U21B2038, U23B2051, 61931008, 62122075 and 61976202, 62206264, 92370102, in part by Youth Innovation Promotion Association CAS, in part by the Strategic Priority Research Program of the Chinese Academy of Sciences, Grant No. XDB0680000, in part by the Innovation Funding of ICT, CAS under Grant No.E000000. 
 
\section*{Impact Statement}

We propose a general multi-modal learning method to deal with the bias toward weak modalities. When the weak modalities are sensitive to a potential group of people in society, it might be helpful to improve the overall fairness of the learning system.

\nocite{langley00}
\bibliography{example_paper}

\begin{thebibliography}{69}
\providecommand{\natexlab}[1]{#1}
\providecommand{\url}[1]{\texttt{#1}}
\expandafter\ifx\csname urlstyle\endcsname\relax
  \providecommand{\doi}[1]{doi: #1}\else
  \providecommand{\doi}{doi: \begingroup \urlstyle{rm}\Url}\fi

\bibitem[Badirli et~al.(2020)Badirli, Liu, Xing, Bhowmik, and Keerthi]{grownet}
Badirli, S., Liu, X., Xing, Z., Bhowmik, A., and Keerthi, S.~S.
\newblock Gradient boosting neural networks: Grownet.
\newblock \emph{ArXiv}, abs/2002.07971, 2020.

\bibitem[Baltrusaitis et~al.(2018)Baltrusaitis, Zadeh, Lim, and Morency]{openface}
Baltrusaitis, T., Zadeh, A., Lim, Y.~C., and Morency, L.-P.
\newblock Openface 2.0: Facial behavior analysis toolkit.
\newblock In \emph{IEEE International Conference on Automatic Face and Gesture Recognition}, pp.\  59--66, 2018.

\bibitem[Baltrušaitis et~al.(2019)Baltrušaitis, Ahuja, and Morency]{mm-survey}
Baltrušaitis, T., Ahuja, C., and Morency, L.-P.
\newblock Multimodal machine learning: A survey and taxonomy.
\newblock \emph{IEEE TPAMI}, 41\penalty0 (2):\penalty0 423--443, 2019.

\bibitem[Cai et~al.(2011)Cai, Nie, Huang, and Kamangar]{jiangdm2c_5}
Cai, X., Nie, F., Huang, H., and Kamangar, F.
\newblock Heterogeneous image feature integration via multi-modal spectral clustering.
\newblock In \emph{CVPR}, pp.\  1977--1984, 2011.

\bibitem[Cao et~al.(2014)Cao, Cooper, Keutmann, Gur, Nenkova, and Verma]{cremad}
Cao, H., Cooper, D.~G., Keutmann, M.~K., Gur, R.~C., Nenkova, A., and Verma, R.
\newblock Crema-d: Crowd-sourced emotional multimodal actors dataset.
\newblock \emph{IEEE Trans. Affective Comput.}, 5\penalty0 (4):\penalty0 377--390, 2014.

\bibitem[Chen et~al.(2022)Chen, Lu, Williamson, Chen, Lipkova, Noor, Shaban, Shady, Williams, Joo, et~al.]{medical_deep_learning}
Chen, R.~J., Lu, M.~Y., Williamson, D.~F., Chen, T.~Y., Lipkova, J., Noor, Z., Shaban, M., Shady, M., Williams, M., Joo, B., et~al.
\newblock Pan-cancer integrative histology-genomic analysis via multimodal deep learning.
\newblock \emph{Cancer Cell}, 40\penalty0 (8):\penalty0 865--878, 2022.

\bibitem[Chen \& Guestrin(2016)Chen and Guestrin]{xgboost}
Chen, T. and Guestrin, C.
\newblock Xgboost: A scalable tree boosting system.
\newblock In \emph{SIGKDD}, pp.\  785--794, 2016.

\bibitem[Chen et~al.(2020)Chen, Lin, Wang, Wu, Qian, Li, and Zeng]{fusion_2}
Chen, X., Lin, K.-Y., Wang, J., Wu, W., Qian, C., Li, H., and Zeng, G.
\newblock Bi-directional cross-modality feature propagation with separation-and-aggregation gate for rgb-d semantic segmentation.
\newblock In \emph{ECCV}, pp.\  561--577, 2020.

\bibitem[Degottex et~al.(2014)Degottex, Kane, Drugman, Raitio, and Scherer]{MOSI-AUDIO}
Degottex, G., Kane, J., Drugman, T., Raitio, T., and Scherer, S.
\newblock Covarep — a collaborative voice analysis repository for speech technologies.
\newblock In \emph{ICASSP}, pp.\  960--964, 2014.

\bibitem[Deng \& Dragotti(2021)Deng and Dragotti]{fusion_4}
Deng, X. and Dragotti, P.~L.
\newblock Deep convolutional neural network for multi-modal image restoration and fusion.
\newblock \emph{IEEE TPAMI}, 43\penalty0 (10):\penalty0 3333--3348, 2021.

\bibitem[Devlin et~al.(2018)Devlin, Chang, Lee, and Toutanova]{bert}
Devlin, J., Chang, M.-W., Lee, K., and Toutanova, K.
\newblock Bert: Pre-training of deep bidirectional transformers for language understanding.
\newblock \emph{arXiv preprint arXiv:1810.04805}, 2018.

\bibitem[Du et~al.(2023)Du, Teng, Li, Liu, Yuan, Wang, Yuan, and Zhao]{UMT}
Du, C., Teng, J., Li, T., Liu, Y., Yuan, T., Wang, Y., Yuan, Y., and Zhao, H.
\newblock On uni-modal feature learning in supervised multi-modal learning.
\newblock In \emph{ICML}, pp.\ ~25, 2023.

\bibitem[Fan et~al.(2023)Fan, Xu, Wang, Wang, and Guo]{PMR_CVPR_23}
Fan, Y., Xu, W., Wang, H., Wang, J., and Guo, S.
\newblock Pmr: Prototypical modal rebalance for multimodal learning.
\newblock In \emph{CVPR}, pp.\  20029--20038, 2023.

\bibitem[Feng et~al.(2022)Feng, Gao, Zhao, Guo, Bagewadi, Bui, Dao, Gangisetty, Guan, Han, et~al.]{shrec}
Feng, Y., Gao, Y., Zhao, X., Guo, Y., Bagewadi, N., Bui, N.-T., Dao, H., Gangisetty, S., Guan, R., Han, X., et~al.
\newblock Shrec’22 track: Open-set 3d object retrieval.
\newblock \emph{Computers \& Graphics}, 107:\penalty0 231--240, 2022.

\bibitem[Freund(1995)]{freund1995boosting}
Freund, Y.
\newblock Boosting a weak learning algorithm by majority.
\newblock \emph{Inf. Comput.}, 121\penalty0 (2):\penalty0 256--285, 1995.

\bibitem[Freund \& Schapire(1997)Freund and Schapire]{freund1997decision}
Freund, Y. and Schapire, R.~E.
\newblock A decision-theoretic generalization of on-line learning and an application to boosting.
\newblock \emph{J. Comput. Syst. Sci.}, 55\penalty0 (1):\penalty0 119--139, 1997.

\bibitem[Freund et~al.(1996)Freund, Schapire, et~al.]{freund1996experiments}
Freund, Y., Schapire, R.~E., et~al.
\newblock Experiments with a new boosting algorithm.
\newblock In \emph{ICML}, pp.\  148--156, 1996.

\bibitem[Friedman(2001)]{gb}
Friedman, J.~H.
\newblock {Greedy function approximation: A gradient boosting machine.}
\newblock \emph{The Annals of Statistics}, 29\penalty0 (5):\penalty0 1189 -- 1232, 2001.

\bibitem[Gallego et~al.(2022)Gallego, Delbruck, Orchard, Bartolozzi, Taba, Censi, Leutenegger, Davison, Conradt, Daniilidis, and Scaramuzza]{event_camera}
Gallego, G., Delbruck, T., Orchard, G., Bartolozzi, C., Taba, B., Censi, A., Leutenegger, S., Davison, A.~J., Conradt, J., Daniilidis, K., and Scaramuzza, D.
\newblock Event-based vision: A survey.
\newblock \emph{IEEE TPAMI}, 44\penalty0 (1):\penalty0 154--180, 2022.

\bibitem[Gao et~al.(2023)Gao, Li, Li, Guo, and Dai]{siqi_vfi}
Gao, Y., Li, S., Li, Y., Guo, Y., and Dai, Q.
\newblock Superfast: 200× video frame interpolation via event camera.
\newblock \emph{IEEE TPAMI}, 45\penalty0 (6):\penalty0 7764--7780, 2023.

\bibitem[Guan et~al.(2018)Guan, Cao, Liang, Cao, and Yang]{dynamic_fusion_2}
Guan, D., Cao, Y., Liang, J., Cao, Y., and Yang, M.~Y.
\newblock Fusion of multispectral data through illumination-aware deep neural networks for pedestrian detection.
\newblock \emph{ArXiv}, abs/1802.09972, 2018.

\bibitem[Han et~al.(2021)Han, Zhang, Fu, and Zhou]{TMC}
Han, Z., Zhang, C., Fu, H., and Zhou, J.~T.
\newblock Trusted multi-view classification.
\newblock In \emph{ICLR}, 2021.

\bibitem[He et~al.(2016)He, Zhang, Ren, and Sun]{resnet}
He, K., Zhang, X., Ren, S., and Sun, J.
\newblock Deep residual learning for image recognition.
\newblock In \emph{CVPR}, 2016.

\bibitem[Hessel \& Lee(2020)Hessel and Lee]{emnlp_cross_modal}
Hessel, J. and Lee, L.
\newblock Does my multimodal model learn cross-modal interactions? it{'}s harder to tell than you might think!
\newblock In \emph{EMNLP}, pp.\  861--877, 2020.

\bibitem[Hu et~al.(2017)Hu, Miyato, Tokui, Matsumoto, and Sugiyama]{jiangdm2c_17}
Hu, W., Miyato, T., Tokui, S., Matsumoto, E., and Sugiyama, M.
\newblock Learning discrete representations via information maximizing self-augmented training.
\newblock In \emph{ICML}, pp.\  1558--1567, 2017.

\bibitem[Huang et~al.(2018)Huang, Ash, Langford, and Schapire]{learning_deep_resnet}
Huang, F., Ash, J., Langford, J., and Schapire, R.
\newblock Learning deep resnet blocks sequentially using boosting theory.
\newblock \emph{ArXiv}, abs/1706.04964, 2018.

\bibitem[Huang et~al.(2022)Huang, Lin, Zhou, Yang, and Huang]{modality_competition}
Huang, Y., Lin, J., Zhou, C., Yang, H., and Huang, L.
\newblock Modality competition: What makes joint training of multi-modal network fail in deep learning? ({P}rovably).
\newblock In \emph{ICML}, pp.\  9226--9259, 2022.

\bibitem[Ivanov \& Prokhorenkova(2021)Ivanov and Prokhorenkova]{bgnn}
Ivanov, S. and Prokhorenkova, L.
\newblock Boost then convolve: Gradient boosting meets graph neural networks.
\newblock In \emph{ICLR}, 2021.

\bibitem[Jiang et~al.(2019)Jiang, Xu, Yang, Cao, and Huang]{jiangdm2c}
Jiang, Y., Xu, Q., Yang, Z., Cao, X., and Huang, Q.
\newblock Dm2c: Deep mixed-modal clustering.
\newblock In \emph{NeurIPS}, pp.\  5880--5890, 2019.

\bibitem[Jiang et~al.(2023{\natexlab{a}})Jiang, Hua, Feng, and Gao]{hsr}
Jiang, Y., Hua, C., Feng, Y., and Gao, Y.
\newblock Hierarchical set-to-set representation for 3-d cross-modal retrieval.
\newblock \emph{IEEE TNNLS}, pp.\  1--13, 2023{\natexlab{a}}.

\bibitem[Jiang et~al.(2023{\natexlab{b}})Jiang, Wang, Li, Zhang, Zhao, and Gao]{yuehang-tmm}
Jiang, Y., Wang, Y., Li, S., Zhang, Y., Zhao, M., and Gao, Y.
\newblock Event-based low-illumination image enhancement.
\newblock \emph{IEEE TMM}, pp.\  1--12, 2023{\natexlab{b}}.

\bibitem[Jing et~al.(2021)Jing, Vahdani, Tan, and Tian]{cmcl}
Jing, L., Vahdani, E., Tan, J., and Tian, Y.
\newblock Cross-modal center loss for 3d cross-modal retrieval.
\newblock In \emph{CVPR}, pp.\  3142--3151, 2021.

\bibitem[Joze et~al.(2020)Joze, Shaban, Iuzzolino, and Koishida]{mmtm}
Joze, H. R.~V., Shaban, A., Iuzzolino, M.~L., and Koishida, K.
\newblock Mmtm: Multimodal transfer module for cnn fusion.
\newblock In \emph{CVPR}, pp.\  13289--13299, 2020.

\bibitem[Kullback \& Leibler(1951)Kullback and Leibler]{kl-divergence}
Kullback, S. and Leibler, R.~A.
\newblock On information and sufficiency.
\newblock \emph{The annals of mathematical statistics}, 22\penalty0 (1):\penalty0 79--86, 1951.

\bibitem[Li et~al.(2022)Li, Qiang, Zheng, Su, Razzak, Wen, and Xiong]{interaction_tkde}
Li, J., Qiang, W., Zheng, C., Su, B., Razzak, F., Wen, J.-R., and Xiong, H.
\newblock Modeling multiple views via implicitly preserving global consistency and local complementarity.
\newblock \emph{IEEE TKDE}, 2022.

\bibitem[Li et~al.(2019)Li, Tang, and Mei]{jinhui-pami-19}
Li, Z., Tang, J., and Mei, T.
\newblock Deep collaborative embedding for social image understanding.
\newblock \emph{IEEE TPAMI}, 41\penalty0 (9):\penalty0 2070--2083, 2019.

\bibitem[Liang et~al.(2023{\natexlab{a}})Liang, Deng, Ma, Zou, Morency, and Salakhutdinov]{mutual_information}
Liang, P.~P., Deng, Z., Ma, M.~Q., Zou, J., Morency, L.-P., and Salakhutdinov, R.
\newblock Factorized contrastive learning: Going beyond multi-view redundancy.
\newblock In \emph{NeurIPS}, 2023{\natexlab{a}}.

\bibitem[Liang et~al.(2023{\natexlab{b}})Liang, Deng, Ma, Zou, Morency, and Salakhutdinov]{factor-contrastive-learning}
Liang, P.~P., Deng, Z., Ma, M.~Q., Zou, J.~Y., Morency, L.-P., and Salakhutdinov, R.
\newblock Factorized contrastive learning: Going beyond multi-view redundancy.
\newblock In \emph{NeurIPS}, pp.\  32971--32998, 2023{\natexlab{b}}.

\bibitem[Liang et~al.(2023{\natexlab{c}})Liang, Lyu, Chhablani, Jain, Deng, Wang, Morency, and Salakhutdinov]{liang2023multiviz}
Liang, P.~P., Lyu, Y., Chhablani, G., Jain, N., Deng, Z., Wang, X., Morency, L.-P., and Salakhutdinov, R.
\newblock Multiviz: Towards visualizing and understanding multimodal models.
\newblock In \emph{ICLR}, 2023{\natexlab{c}}.

\bibitem[McFee et~al.(2015)McFee, Raffel, Liang, Ellis, McVicar, Battenberg, and Nieto]{librosa}
McFee, B., Raffel, C., Liang, D., Ellis, D.~P., McVicar, M., Battenberg, E., and Nieto, O.
\newblock librosa: Audio and music signal analysis in python.
\newblock In \emph{SciPy}, pp.\  18--24, 2015.

\bibitem[Nagrani et~al.(2021)Nagrani, Yang, Arnab, Jansen, Schmid, and Sun]{intermediate_fusion2}
Nagrani, A., Yang, S., Arnab, A., Jansen, A., Schmid, C., and Sun, C.
\newblock Attention bottlenecks for multimodal fusion.
\newblock In \emph{NeurIPS}, pp.\  14200--14213, 2021.

\bibitem[Paszke et~al.(2017)Paszke, Gross, Chintala, Chanan, Yang, DeVito, Lin, Desmaison, Antiga, and Lerer]{paszke2017automatic}
Paszke, A., Gross, S., Chintala, S., Chanan, G., Yang, E., DeVito, Z., Lin, Z., Desmaison, A., Antiga, L., and Lerer, A.
\newblock Automatic differentiation in pytorch.
\newblock 2017.

\bibitem[Peng et~al.(2022)Peng, Wei, Deng, Wang, and Hu]{OGM_GE_2022_CVPR}
Peng, X., Wei, Y., Deng, A., Wang, D., and Hu, D.
\newblock Balanced multimodal learning via on-the-fly gradient modulation.
\newblock In \emph{CVPR}, pp.\  8238--8247, 2022.

\bibitem[Robbins \& Monro(1951)Robbins and Monro]{gradient_descent}
Robbins, H. and Monro, S.
\newblock A stochastic approximation method.
\newblock \emph{The annals of mathematical statistics}, pp.\  400--407, 1951.

\bibitem[Ruan et~al.(2021)Ruan, Ji, Yan, Zhu, Zhao, Yang, Gao, Zou, and Dai]{hypergraph}
Ruan, D., Ji, S., Yan, C., Zhu, J., Zhao, X., Yang, Y., Gao, Y., Zou, C., and Dai, Q.
\newblock Exploring complex and heterogeneous correlations on hypergraph for the prediction of drug-target interactions.
\newblock \emph{Patterns}, 2\penalty0 (12), 2021.

\bibitem[Seichter et~al.(2021)Seichter, Köhler, Lewandowski, Wengefeld, and Gross]{intermediate_fusion1}
Seichter, D., Köhler, M., Lewandowski, B., Wengefeld, T., and Gross, H.-M.
\newblock Efficient rgb-d semantic segmentation for indoor scene analysis.
\newblock \emph{ArXiv}, abs/2011.06961, 2021.

\bibitem[Shahroudy et~al.(2017)Shahroudy, Ng, Gong, and Wang]{fusion_1}
Shahroudy, A., Ng, T.-T., Gong, Y., and Wang, G.
\newblock Deep multimodal feature analysis for action recognition in rgb+ d videos.
\newblock \emph{IEEE TPAMI}, 40\penalty0 (5):\penalty0 1045--1058, 2017.

\bibitem[Shalev-Shwartz(2014)]{shalev2014selfieboost}
Shalev-Shwartz, S.
\newblock Selfieboost: A boosting algorithm for deep learning.
\newblock \emph{ArXiv}, abs/1411.3436, 2014.

\bibitem[Shao et~al.(2023)Shao, Li, Zhou, Chen, Zhu, and Yao]{shao2023identity}
Shao, Z., Li, F., Zhou, Y., Chen, H., Zhu, H., and Yao, R.
\newblock Identity-invariant representation and transformer-style relation for micro-expression recognition.
\newblock \emph{Applied Intelligence}, 53\penalty0 (17):\penalty0 19860--19871, 2023.

\bibitem[Shao et~al.(2024)Shao, Zhou, Li, Zhu, and Liu]{shao2024joint}
Shao, Z., Zhou, Y., Li, F., Zhu, H., and Liu, B.
\newblock Joint facial action unit recognition and self-supervised optical flow estimation.
\newblock \emph{Pattern Recognition Letters}, 181:\penalty0 70--76, 2024.

\bibitem[Sun et~al.(2019)Sun, Zhu, and Lin]{adagcn}
Sun, K., Zhu, Z., and Lin, Z.
\newblock Adagcn: Adaboosting graph convolutional networks into deep models.
\newblock \emph{ArXiv}, abs/1908.05081, 2019.

\bibitem[Tang et~al.(2017)Tang, Shu, Qi, Li, Wang, Yan, and Jain]{jinhui-pami-17}
Tang, J., Shu, X., Qi, G.-J., Li, Z., Wang, M., Yan, S., and Jain, R.
\newblock Tri-clustered tensor completion for social-aware image tag refinement.
\newblock \emph{IEEE TPAMI}, 39\penalty0 (8):\penalty0 1662--1674, 2017.

\bibitem[Tian et~al.(2018)Tian, Shi, Li, Duan, and Xu]{ave}
Tian, Y., Shi, J., Li, B., Duan, Z., and Xu, C.
\newblock Audio-visual event localization in unconstrained videos.
\newblock In \emph{ECCV}, pp.\  247--263, 2018.

\bibitem[Van~der Maaten \& Hinton(2008)Van~der Maaten and Hinton]{tsne}
Van~der Maaten, L. and Hinton, G.
\newblock Visualizing data using t-sne.
\newblock \emph{JMLR}, 9\penalty0 (11), 2008.

\bibitem[Wan et~al.(2023)Wan, Mao, Zhang, and Dai]{optical_flow}
Wan, Z., Mao, Y., Zhang, J., and Dai, Y.
\newblock Rpeflow: Multimodal fusion of rgb-pointcloud-event for joint optical flow and scene flow estimation.
\newblock In \emph{ICCV}, pp.\  10030--10040, 2023.

\bibitem[Wang et~al.(2020{\natexlab{a}})Wang, Tran, and Feiszli]{g-blending}
Wang, W., Tran, D., and Feiszli, M.
\newblock What makes training multi-modal classification networks hard?
\newblock In \emph{CVPR}, pp.\  12692--12702, 2020{\natexlab{a}}.

\bibitem[Wang et~al.(2020{\natexlab{b}})Wang, Huang, Sun, Xu, Rong, and Huang]{fusion_3}
Wang, Y., Huang, W., Sun, F., Xu, T., Rong, Y., and Huang, J.
\newblock Deep multimodal fusion by channel exchanging.
\newblock In \emph{NeurIPS}, pp.\  4835--4845, 2020{\natexlab{b}}.

\bibitem[Wang et~al.(2024)Wang, Zhang, Guo, Zhao, and Jiang]{yuehang-spl}
Wang, Y., Zhang, Y., Guo, Q., Zhao, M., and Jiang, Y.
\newblock Rnve: A real nighttime vision enhancement benchmark and dual-stream fusion network.
\newblock \emph{IEEE Signal Process Lett.}, 31:\penalty0 131--135, 2024.

\bibitem[Wei et~al.(2022)Wei, Hu, Tian, and Li]{wei2022-ml-survey}
Wei, Y., Hu, D., Tian, Y., and Li, X.
\newblock Learning in audio-visual context: A review, analysis, and new perspective, 2022.

\bibitem[Williams et~al.(2018)Williams, Kleinegesse, Comanescu, and Radu]{emotion-dnn}
Williams, J., Kleinegesse, S., Comanescu, R., and Radu, O.
\newblock Recognizing emotions in video using multimodal dnn feature fusion.
\newblock In \emph{Proceedings of Grand Challenge and Workshop on Human Multimodal Language}, 2018.
\newblock \doi{10.18653/v1/W18-3302}.

\bibitem[Wu et~al.(2022)Wu, Jastrzebski, Cho, and Geras]{greedy}
Wu, N., Jastrzebski, S., Cho, K., and Geras, K.~J.
\newblock Characterizing and overcoming the greedy nature of learning in multi-modal deep neural networks.
\newblock In \emph{ICML}, pp.\  24043--24055, 2022.

\bibitem[Wu et~al.(2015)Wu, Song, Khosla, Yu, Zhang, Tang, and Xiao]{mn40}
Wu, Z., Song, S., Khosla, A., Yu, F., Zhang, L., Tang, X., and Xiao, J.
\newblock 3d shapenets: A deep representation for volumetric shapes.
\newblock In \emph{CVPR}, pp.\  1912--1920, 2015.

\bibitem[Yang et~al.(2022)Yang, Wang, Duan, Chen, Hou, Jin, and Zhu]{avqa}
Yang, P., Wang, X., Duan, X., Chen, H., Hou, R., Jin, C., and Zhu, W.
\newblock Avqa: A dataset for audio-visual question answering on videos.
\newblock In \emph{ACM MM}, pp.\  3480–3491, 2022.

\bibitem[Yu et~al.(2020)Yu, Xu, Meng, Zhu, Ma, Wu, Zou, and Yang]{ch-sims}
Yu, W., Xu, H., Meng, F., Zhu, Y., Ma, Y., Wu, J., Zou, J., and Yang, K.
\newblock Ch-sims: A chinese multimodal sentiment analysis dataset with fine-grained annotation of modality.
\newblock In \emph{Annual Meeting of the Association for Computational Linguistics}, 2020.

\bibitem[Zadeh et~al.(2016)Zadeh, Zellers, Pincus, and Morency]{mosi}
Zadeh, A., Zellers, R., Pincus, E., and Morency, L.-P.
\newblock Mosi: Multimodal corpus of sentiment intensity and subjectivity analysis in online opinion videos.
\newblock \emph{ArXiv}, abs/1606.06259, 2016.

\bibitem[Zadeh et~al.(2018)Zadeh, Liang, Poria, Cambria, and Morency]{mosei}
Zadeh, A., Liang, P.~P., Poria, S., Cambria, E., and Morency, L.-P.
\newblock Multimodal language analysis in the wild: Cmu-mosei dataset and interpretable dynamic fusion graph.
\newblock In \emph{Annual Meeting of the Association for Computational Linguistics}, 2018.

\bibitem[Zhang et~al.(2020)Zhang, Zhang, Tang, Hua, and Sun]{jinhui-nips-20}
Zhang, D., Zhang, H., Tang, J., Hua, X.-S., and Sun, Q.
\newblock Causal intervention for weakly-supervised semantic segmentation.
\newblock In Larochelle, H., Ranzato, M., Hadsell, R., Balcan, M., and Lin, H. (eds.), \emph{NeurIPS}, volume~33, pp.\  655--666, 2020.

\bibitem[Zhang et~al.(2023)Zhang, Wu, Zhang, Hu, Fu, Zhou, and Peng]{dynamic_fusion}
Zhang, Q., Wu, H., Zhang, C., Hu, Q., Fu, H., Zhou, J.~T., and Peng, X.
\newblock Provable dynamic fusion for low-quality multimodal data.
\newblock In \emph{ICML}, pp.\ ~17, 2023.

\bibitem[Zhang et~al.(2024)Zhang, Wei, Han, Fu, Peng, Deng, Hu, Xu, Wen, Hu, and Zhang]{zhang-ml-survey}
Zhang, Q., Wei, Y., Han, Z., Fu, H., Peng, X., Deng, C., Hu, Q., Xu, C., Wen, J., Hu, D., and Zhang, C.
\newblock Multimodal fusion on low-quality data: A comprehensive survey, 2024.

\end{thebibliography}
\bibliographystyle{icml2024}


\newpage
\appendix
\onecolumn
\definecolor{blue}{RGB}{0,20,115}
\textcolor{white}{dasdsa}
\section*{\textcolor{blue}{\Large{Contents}}}
\setcounter{tocdepth}{2}  
\titlecontents{section}[0em]{\color{blue}\bfseries}{\thecontentslabel. }{}{\hfill\contentspage}

\titlecontents{subsection}[1.5em]{\color{blue}}{\thecontentslabel. }{}{\titlerule*[0.75em]{.}\contentspage} 

\startcontents[sections]
\printcontents[sections]{l}{1}{}
\newpage

\section{Prior Arts}
\label{sec:related}
In this section, we briefly review the closely related studies along with our main topic.
\subsection{Multi-modal Learning}

Recent decades have witnessed the development of multi-modal learning research which covers various fields like cross-modal retrieval \cite{hsr,shrec}, video frame interpolation \cite{siqi_vfi}, image reconstruction \cite{yuehang-spl,yuehang-tmm}, visual question answering \cite{avqa}, and clustering \cite{jiangdm2c,jiangdm2c_17}. Intuitively, multi-modal models integrate information from multiple sensors to outperform their uni-modal counterparts. For example, event cameras as new vision sensors can compensate for the shortcomings of standard cameras in the face of abnormal light conditions or challenging high-speed scenarios \cite{event_camera}. These examples underscore the effectiveness of multi-modal approaches in addressing specific challenges and highlight the advantages arising from the fusion of diverse sensor modalities. 

Numerous studies \cite{liang2023multiviz, UMT,liang2023multiviz,emnlp_cross_modal,dynamic_fusion,interaction_tkde,mutual_information} mainly concentrate on integrating modality-specific features into a shared representation for diverse tasks. Employed fusion methods encompass early/intermediate fusion \cite{intermediate_fusion1, intermediate_fusion2} as well as late fusion \cite{OGM_GE_2022_CVPR, PMR_CVPR_23,UMT,g-blending}. Recent intermediate fusion methods utilize attention mechanisms that connect multi-modal signals during the modality-specific feature learning stage \cite{intermediate_fusion2}. While intermediate fusion may enhance representation learning, late fusion consistently stands out as the most prevalent and widely used approach, owing to its interpretability and practicality. Evolving from the naive late-fusion method, more methods are using dynamic fusion \cite{dynamic_fusion_2,dynamic_fusion} approaches to unleash the value of each modality and reduce the impact of low-quality multi-modal data.

\subsection{Balanced Multi-modal Learning}

However, recent theoretical evidence \cite{modality_competition} illustrated that current paradigms of multi-modal learning encounter \textbf{\textit{Modality Competition.}} Such a problem occurs when the objective for different modalities is optimized synchronously. In optimization, the modality with faster convergence dominates the learning process. Therefore, the learning parameters of other modalities can not be updated in a timely and effective manner. It will limit the optimization of the uni-modal branch and cannot fully exploit the information of the uni-modal, becoming a
bottleneck in the performance of multi-modal learning.

To fill this gap, several studies \cite{g-blending,OGM_GE_2022_CVPR,UMT,PMR_CVPR_23} are proposed to balance the optimization process across different modality learners and promote the uni-modal learning. G-Blending \cite{g-blending} incorporates uni-modal classifiers with extra supervised signals in multi-modal learning to effectively blend modalities. OGM-GE \cite{OGM_GE_2022_CVPR} focuses on suppressing the dominant modality and assisting the training of others through adaptive gradient modulations. 
PMR \cite{PMR_CVPR_23} employs the prototypical cross-entropy loss to accelerate the learning process of the weak modality. Additionally, UMT \cite{UMT} distills knowledge from well-trained uni-modal models in multi-modal learning, which can effectively benefit from uni-modal learning. In general, the majority of prior studies adopt a synchronous learning paradigm.

\subsection{Boosting}

Boosting is a commonly used learning approach in machine learning \cite{gb,freund1995boosting,gb,freund1996experiments,freund1997decision}. It enhances the performance of a basic learner by combining multiple weaker learners. In each iteration of boosting, the weaker learner focuses on the residual between the truth and its estimation. Decision trees are the most common weak learners that are used in boosting frameworks. Popular boosting algorithms include AdaBoost \cite{freund1997decision}, GBDT\cite{gb}, and XGBoost \cite{xgboost}.

Inspired by the success of boosting in machine learning, boosting has recently received research attention in the deep learning community. Unlike traditional methods to construct ensembles of learners, SelfieBoost \cite{shalev2014selfieboost} boosts the accuracy of a single network while discarding intermediate learners. \cite{learning_deep_resnet} builds a ResNet-style architecture based on multi-channel telescoping sum boosting theory. AdaGCN \cite{adagcn} interprets a multi-scale graph convolutional network as an ensemble model and trains it using AdaBoost. BGNN \cite{bgnn} combines the GBDT and GNN by iteratively adding new trees that fit the gradient updates of GNN.

\section{Proof of Theorem \ref{thm:main}}
\label{app:proof}

\begin{rthm1}
When the reconcilement regularization satisfies, 
\begin{equation*}
    \lambda\cdot\nabla_{\phi_k}\mathbb{D}_{s}\left(\Phi_{M/k}(x_i),\phi_k (\vartheta_k;m_i^k)\right) = \nabla_{\phi_k}\ell\left(\phi_k (\vartheta_k;m_i^k),y_i\right) -  \nabla_{\phi_k}\ell\left(\phi_k (\vartheta_k;m_i^k),-\nabla_{\Phi_{M/k}}\ell(\Phi_{M/k}(x_i),y_i)\right)
\end{equation*}

It leads to equivalent optimization goals:
\begin{equation*}
\begin{aligned}
    \nabla_{\vartheta_k}{\tilde{\mathcal{L}}}_s\left(\phi_k(m^k),{y}\right) \iff \nabla_{\vartheta_k}{\mathcal{L}}\left(\phi_{k}(m^k), {-\nabla_{\Phi_{M/k}} \ell(\Phi_{M/k}(x), y)}\right)
\end{aligned}
\end{equation*}
\end{rthm1}

\begin{proof}
By using the right-hand side as the objective for gradient boosting, the $k$-th modality learner's parameters update as follows:
\begin{align}
\vartheta_k^{t+1} &=  \vartheta_k^{t} - \eta\cdot\nabla_{\vartheta_k^{t}}\frac{1}{N}\sum_{i=1}^{N} \ell\left(\phi_k\left(\vartheta_k^t;m_i^k\right),-\nabla_{\Phi_{M/k}}\ell\left(\Phi_{M/k}(x_i),y_i\right)\right) \\
 &= \vartheta_k^{t} - \eta_1\cdot\frac{1}{N}\sum_{i=1}^{N} \left(\frac{\partial \phi_k(\vartheta_k^t;m_i^k)}{\partial \vartheta_k^{t}} \right)^{T} \nabla_{\phi_k}\ell\left(\phi_k \left(\vartheta_k;m_i^k\right),-\nabla_{\Phi_{M/k}}\ell\left(\Phi_{M/k}(x_i),y_i\right)\right)        
\end{align}
If the left-hand side is the optimization strategy, then the objective becomes:
\begin{align}
    {\tilde{\mathcal{L}}}_s(\phi_k(m^k),{y})  = \frac{1}{N}\sum_{i=1}^{N}\left [\ell\left(\phi_k\left(\vartheta_k;m_i^k\right),y_i\right) - \lambda\cdot \underbrace{ \mathbb{D}_{s}\left(\Phi_{M/k}(x_i),\phi_k \left(\vartheta_k;m_i^k\right)\right)}_{\text{reconcilement regularization term}} \right ]
\end{align}
Through gradient optimization, we update the $k$-th modality learner's parameters:
\begin{align}
\vartheta_k^{t+1} &= \vartheta_k^{t} - \eta \cdot\nabla_{\vartheta_k^{t}}{\tilde{\mathcal{L}}}_s\left(\phi_k(m^k),{y}\right) \\
&=  \vartheta_k^{t} - \eta \cdot \frac{1}{N}\sum_{i=1}^{N}\left(\frac{\partial \phi_k\left(\vartheta_k^t;m_i^k\right)}{\partial \vartheta_k^{t}} \right)^{T}\cdot\left[\nabla_{\phi_k}\ell\left(\phi_k (\vartheta_k;m_i^k),y_i\right) - \lambda\cdot\nabla_{\phi_k}\mathbb{D}_{s}\left(\Phi_{M/k}(x_i),\phi_k \left(\vartheta_k;m_i^k\right)\right) \right] \\
&= \vartheta_k^{t} - \eta \cdot\frac{1}{N}\sum_{i=1}^{N} \left(\frac{\partial \phi_k\left(\vartheta_k^t;m_i^k\right)}{\partial \vartheta_k^{t}} \right)^{T} \nabla_{\phi_k}\ell\left(\phi_k \left(\vartheta_k;m_i^k\right),-\nabla_{\Phi_{M/k}}\ell\left(\Phi_{M/k}\left(x_i\right),y_i\right)\right)
\end{align}
Thus, we conclude that 
\begin{equation}
       \nabla_{\vartheta_k}{\tilde{\mathcal{L}}}_s\left(\phi_k(m^k),{ y}\right) \iff \nabla_{\vartheta_k}{\mathcal{L}}\left(\phi_{k}(m^k), {-\nabla_{\Phi_{M/k}} \ell\left(\Phi_{M/k}(x), y\right)}\right) 
\end{equation}

\end{proof}

Here, to better understand the generality of our method and theorem, we will consider the case where the optimization loss function is Cross Entropy loss. CE loss is widely used for problems like classification, retrieval, and contrastive learning. As a corollary of the theorem we have:

\begin{gthm1}
Let the reconcilement regularization be a KL divergence \cite{kl-divergence} function:
\begin{align*}
{\mathbb{D}_{s}\left(\Phi_{M/k}\left(x_i\right),\phi_k \left(\vartheta_k;m_i^k\right)\right) = \mathbb{D}_{KL,s}\left(\Phi_{M/k}\left(x_i\right)\|\phi_k \left(\vartheta_k;m_i^k\right)\right)}   
\end{align*}
Then, 
\begin{equation*}
\begin{aligned}
\nabla_{\vartheta_k}{\tilde{\mathcal{L}}}_s\left(\phi_k\left(m^k\right),{y}\right) \iff \nabla_{\vartheta_k}{\mathcal{L}}\left(\phi_{k}(m^k), {-\nabla_{\Phi_{M/k}} \ell\left(\Phi_{M/k}(x), y\right)}\right)
\end{aligned}
\end{equation*}
where $\ell$ is the CE loss.
\end{gthm1}
\begin{proof}

By using the right-hand side as the objective for gradient boosting, the $k$-th modality learner's parameters can be updated as:
\begin{align}
    \vartheta_k^{t+1} = \vartheta_k^{t} - \eta\cdot\nabla_{\vartheta_k^{t}}\frac{1}{N}\sum_{i=1}^{N} \ell(\phi_k(\vartheta_k^t;m_i^k),\Tilde{y}_i) 
\end{align}
The pseudo-label $\Tilde{y}_i$ is
\begin{align}
     \Tilde{y}_i = - \frac{\partial\ell(\Phi_{M/k}(x_i),y_i)}{\partial\Phi_{M/k}(x_i)}  = y_i- \sigma_i
\end{align}
$\sigma_{i}\in \mathbb{R}^Y$ means the prediction score for $i$-th sample. 

Therefore, we have
\begin{align}
       \vartheta_k^{t+1} =  \vartheta_k^{t} -  \eta\cdot\nabla_{\vartheta_k^{t}} \frac{1}{N}\sum_{i=1}^{N} \sum_{j=1}^{Y}(\sigma_{i,j} - c_{i,j})\cdot \log(\rho_{i,j}^{k,t})
\end{align}
Here, $\rho_{i}^{k}$ is the prediction of the $k$-th modality learner on the $i$-th sample.

The left-hand objective is
\begin{align}
    {\tilde{\mathcal{L}}}_s(\phi_k(m^k),{ y}) & = \frac{1}{N}\sum_{i=1}^{N}\left [\ell\left(\phi_k\left(\vartheta_k;m_i^k\right),y_i\right) - \lambda\cdot \underbrace{\mathbb{D}_{KL,s}\left(\Phi_{M/k}\left(x_i\right)\|\phi_k\left(\vartheta_k;m_i^k\right)\right)}_{\text{KL-based reconcilement regularization}} \right ] \\
    &=  \frac{1}{N}\sum_{i=1}^{N}\left [ - \sum_{j=1}^{Y}c_{i,j}\log\left(\rho^k_{i,j}\right) - \lambda \cdot \sum_{j=1}^{Y} \sigma_{i,j}\ln{\frac{\sigma_{i,j}}{\rho^k_{i,j}}} \right ] \\
    & = \frac{1}{N}\sum_{i=1}^{N} \left[-\sum_{j=1}^{Y}c_{i,j}\log\left(\rho^k_{i,j}\right) + \lambda \cdot \sum_{j=1}^{Y} \sigma_{i,j}\ln{{\rho^k_{i,j}}} -\lambda\cdot \sum_{j=1}^{Y} \sigma_{i,j}\ln{{\sigma_{i,j}}} \right]
\end{align}

Through gradient optimization, in $t$-th iteration, the parameters of the $k$-th modality learner can be updated as:
\begin{align}
      \vartheta_k^{t+1} =& \vartheta_k^{t} - \eta\cdot\nabla_{\vartheta_k^{t}}{\tilde{\mathcal{L}}}_s(\phi_k(m^k),{ y}) \\
    =& \vartheta_k^t - \eta \cdot \nabla_{\vartheta_k^{t}} \frac{1}{N} \sum_{i=1}^{N} \sum_{j=1}^{Y}\left(\lambda\cdot\sigma_{i,j}\ln\rho_{i,j}^k - c_{i,j}\log(\rho_{i,j}^k)\right)
\end{align}

Thus, with specific $\lambda$, we can reach the conclusion.
\end{proof}

\section{Additional Experiment Setting}
\label{app:exp_setting}
In this section, we elaborate on the setup of the main experiment, including dataset description, several state-of-the-art baselines, and implementation details.

\subsection{Dataset Description}
\label{app:dataset}
We perform empirical studies on six public benchmark datasets, including:
\begin{itemize}
    \item \textbf{AVE}\footnote{\url{https://sites.google.com/view/audiovisualresearch}} \cite{ave}. The AVE dataset is designed for audio-visual event localization. The dataset contains $4143$ videos covering $28$ event categories and videos in AVE are temporally labeled with audio-visual event boundaries. Each video contains at least one $2$s long audio-visual event. The dataset covers a wide range of audiovisual events from different domains, such as, human activities, animal activities, music performances, and vehicle sounds. All videos are collected from YouTube. The training and testing split of the dataset follows \cite{ave}.
    \item \textbf{CREMA-D}\footnote{\url{https://github.com/CheyneyComputerScience/CREMA-D}} \cite{cremad}. The CREMA-D dataset  is an audio-visual video dataset for speech emotion recognition, which consists of $7442$ original clips of $2$-$3$ seconds from 91 actors speaking several short words. It comprises six different emotions: anger, disgust, fear, happy, neutral, and sad. Categorical emotion labels were collected using crowd-sourcing from $2443$ raters. The training and testing split of the dataset follows the split \cite{cremad}.

    \item  \textbf{ModelNet40}\footnote{\url{https://modelnet.cs.princeton.edu/}} \cite{mn40}. The ModelNet40 is from a large-scale $3$-D CAD model dataset ModelNet for object classification. ModelNet40 is a subset of ModelNet, which contains $40$ popular object categories. We use the front view and the rear view to classify the $3$-D object, following \cite{greedy} and \cite{UMT}. The dataset split for training and testing follows the standard protocol as described in \cite{mn40}.
    
    \item \textbf{MOSI}\footnote{\url{https://drive.google.com/drive/folders/1A2S4pqCHryGmiqnNSPLv7rEg63WvjCSk?usp=sharing}\label{f1}} \cite{mosi}. The CMU-MOSI dataset is one of the most popular benchmark datasets for multi-modal sentiment analysis (MSA). It comprises $2199$ short monologue video clips taken from 93 Youtube movie review video. Human annotators label each sample with a sentiment score from $-3$ (strongly negative) to $3$ (strongly positive). We view this as a three classification problem, with the categories being negative, neutral, and positive. The training and testing split of the dataset follows the split \cite{mosi}.
    
    \item \textbf{MOSEI}\footref{f1} \cite{mosei}. The CMU-MOSEI dataset expands its data with a higher number of utterances, greater variety in samples, speakers, and topics over CMU-MOSI.The dataset contains $23453$ annotated video utterances, from $5000$ videos, $1000$ distinct speakers and $250$ different topics. The training and testing split of the dataset follows the split \cite{mosei}.
    \item \textbf{CH-SIMS}\footref{f1} \cite{ch-sims}. The SIMS dataset is a Chinese MSA benchmark with fine-grained annotations of modality. The dataset consists of $2281$ refined video segments collected from different movies, TV serials, and variety shows with spontaneous expressions, various head poses, occlusions, and illuminations. Human annotators label each sample with a sentiment score from $-1$ (strongly negative) to $1$ (strongly positive). We treat this as a three classification problem, with the categories being negative, neutral, and positive. The training and testing split of the dataset follows the split \cite{ch-sims}.
\end{itemize}
To summarize, the overall statistical information is included in Tab.\ref{tab:dataset}.
\begin{table}[htbp]
  \centering
  \caption{The statistics of all datasets used in the experiments.}
  \vskip 0.1in
    \begin{tabular}{cccccccc}
    \toprule
    \multirow{2}[4]{*}{Dataset} & \multirow{2}[4]{*}{Task} & \multirow{2}[4]{*}{\# Train} & \multirow{2}[4]{*}{\# Test} &  \multirow{2}[4]{*}{\# Category} & \multicolumn{3}{c}{Modality} \\
\cmidrule{6-8}          &       &       &         &       & Audio & Visual & Text \\
    \midrule
    CREMAD & Speech emotion recognition & $6698$  & $744$     & $6$     &   \checkmark    &   \checkmark    & \XSolidBrush \\
    AVE   & Event localization &   $3339$    &      $402$       & $28$    &    \checkmark   &     \checkmark  & \XSolidBrush \\
    ModelNet40 & Object classification & $9438$  & $2468$         & $40$    &   \XSolidBrush    &   \checkmark    & \XSolidBrush \\
    MOSEI & Emotion recognition & $16327$ & $4659$       & $3$     &  \checkmark     &  \checkmark     &  \checkmark \\
    MOSI  & Emotion recognition &  $1284$     &  $686$           & $3$     &  \checkmark     &  \checkmark     & \checkmark \\
    SIMS  & Emotion recognition &   $1368$    & $457$           & $3$     &   \checkmark    &  \checkmark     &  \checkmark \\
    \bottomrule
    \end{tabular}%
  \label{tab:dataset}%
\end{table}%

\subsection{Competitors}
\label{app:comp}
We compare the performance of our proposed method with several state-of-the-art baselines, including:
\begin{itemize}
    \item \textbf{G-Blending} \cite{g-blending}  proposes Gradient Blending to obtain an optimal blending of modalities based on their over-fitting behaviors.
    \item \textbf{OGM-GE}\footnote{\url{https://github.com/GeWu-Lab/OGM-GE_CVPR2022}} \cite{OGM_GE_2022_CVPR} proposes on-the-fly gradient modulation to adaptively control the optimization of each modality, via monitoring the discrepancy of their contribution towards the learning objective.
    \item \textbf{PMR}\footnote{\url{https://github.com/fanyunfeng-bit/Modal-Imbalance-PMR}} \cite{PMR_CVPR_23} proposes the prototypical modal rebalance strategy to address the modality imbalance problem, accelerating the slow modality with prototypical cross entropy loss and reducing the inhibition from dominant modality with prototypical entropy regularization term.
    \item \textbf{UME}\footnote{\url{https://openreview.net/forum?id=mb7VM83DkyC}\label{ume}} \cite{UMT} weights the predictions of well-trained uni-modal model directly.
    \item \textbf{UMT}\footref{ume} \cite{UMT} distills the well-trained uni-modal features to the corresponding parts of multi-modal late-fusion models and fusion the multi-modal features to obtain the final score.
\end{itemize}

\subsection{Implementation Details}
\label{sec:implement_details}
\subsubsection{Network architecture} 

With respect to AVE, CREMA-D, and ModelNet40 datasets, the ResNet18 \cite{resnet} is adopted as the backbone. \textbf{For AVE,} $3$ frames of size $224\times 224\times 3$ are uniformly sampled from each 10-second clip as visual input and the whole audio data is transformed into a spectrogram of size $257\times 1004$ by librosa\footnote{\url{https://librosa.org/}} using a window with a length of $512$ and overlap of $353$. \textbf{For CREMA-D,} $1$ frame of size $224\times 224 \times 3$ is extracted from each video clip, and audio data is transformed into a spectrogram of size $257\times 299$ with a length of $512$ and overlap of $353$. \textbf{In ModelNet40,} we resize the input front and rear views of a 3D object and CenterCrop it to $224\times 224 \times 3$. To make the model compatible with the different data modalities mentioned above, we modify the input channel of ResNet-18 while keeping the remaining parts unchanged. Specifically, it takes the images as inputs and generates $512$ dimension features, and takes the audio as inputs and outputs $512$ dimension features, respectively. Then, a fully connected layer is established on top of the backbone model to make modality-specific predictions. Finally, multi-modal predictions are merged to obtain the final score.

For MOSEI, MOSI, and SIMS datasets, we conduct experiments with fully customized multimodal features extracted by the MMSA-FET\footnote{\url{https://github.com/thuiar/MMSA-FET}} toolkit. The language features are extracted from pre-trained Bert\cite{bert} and the pre-trained feature dimensions are $768$ for all three datasets. Both MOSI and MOSEI use Facet\footnote{\url{https://imotions.com/products/imotions-lab/}} and SIMS uses the MultiComp OpenFace2.0 toolkit\cite{openface} to extract facial expression features. The pre-trained visual feature dimensions are $20$ for MOSI, $35$ for MOSEI, and $709$ for SIMS. Both MOSI and MOSEI extract acoustic features from COVAREP\cite{MOSI-AUDIO} and SIMS uses LibROSA\cite{librosa} speech toolkit with default parameters to extract acoustic features at $22050$Hz. The pre-trained audio feature dimensions are $74$ for MOSEI, $5$ for MOSI, and $33$ for SIMS. For these three datasets, we feed the pre-trained features into modality-specific backbones to extract the latent feature, with the hidden dimension set to $128$. Following \cite{emotion-dnn}, the AudioNet and VisualNet are composed of three fully connected layers and the TextNet uses LSTM to capture long-distance dependencies in a text sequence. Then, a fully connected layer is established on top of the backbone model to make modality-specific predictions. Finally, multi-modal predictions are merged to obtain the final score.

\subsubsection{Training details}

All experiments are conducted on a Ubuntu 20.04 LTS server equipped with Intel(R) Xeon(R) Gold 5218 CPU@2.30GHz and RTX 3090 GPUs, and we implement all algorithms with PyTorch \cite{paszke2017automatic}. We adopt SGD \cite{gradient_descent} as the optimizer and set the same learning rate in the alternating-boosting stage and rectification stage. The learning rate is $0.01$ initially and multiplies $0.1$ every $30$ stages for the CREMA-D dataset, while multiplies $0.5$ after $40$ stages for the AVE dataset. For MOSEI, MOSI, CH-SIMS, and ModelNet40, the learning rate is $0.01$ and remains constant. In one alternating-boosting stage, we will pick one modality learner to update and this modality learner will experience $T1$ epochs. Then, we will step into global rectification stage and the model will experience $T2$ epochs. $T1$ and $T2$ will vary depending on the datasets. In AVE, CREMA-D, ModelNet40, MOSEI, MOSI and SIMS, $T1$ is $4$, $4$, $4$, $1$, $1$, $1$ and $T2$ is $4$, $4$, $4$, $1$, $1$, $1$ respectively.

\section{Additional Experiment Analysis}
\label{app:exp_analysis}
In this section, we provide additional experimental results and analysis to further support the conclusions in the main text.

\subsection{Performance on Retrieval Task}
\label{sec:app_retrieval}
To further demonstrate ReconBoost's adaptability in broader contexts, we apply it to the retrieval task, a crucial area within computer vision. For this purpose, we employ modality-specific pre-trained encoders to obtain latent features from each modality. For modality-specific retrieval, we utilize the respective latent features, whereas, for holistic retrieval, we combine all latent features to make predictions. Cosine similarity served as the metric for our retrieval scores. We assess its performance using the Mean Average Precision (MAP) metric on the CREMA-D and AVE datasets, as detailed in the subsequent Tab.\ref{tab:retrieval}. 

\begin{table}[]
  \centering
  \renewcommand\arraystretch{1.1}
  \caption{Performance comparisons on the AVE and CREMA-D datasets in terms of mAP(\%).}
  \vskip 0.1in
    \begin{tabular}{c|cccccc}
    \toprule
    \multicolumn{1}{c}{\multirow{2}[4]{*}{\textbf{Method}}} & \multicolumn{3}{c}{\textbf{AVE}} & \multicolumn{3}{c}{\textbf{CREMA-D}} \\
\cmidrule{2-7}    \multicolumn{1}{c}{} & MAP   & Audio MAP & Visual MAP & MAP   & Audio MAP & Visual MAP \\
    \midrule
    Concat Fusion & \cellcolor[rgb]{ 1,  .992,  .988}35.25  & \cellcolor[rgb]{ 1,  .91,  .859}37.23  & 18.82  & 36.43  & 34.71  & 20.08  \\
    OGM-GE & \cellcolor[rgb]{ 1,  .953,  .922}36.92  & \cellcolor[rgb]{ 1,  .953,  .925}35.43  & \cellcolor[rgb]{ 1,  .961,  .933}20.04  & \cellcolor[rgb]{ 1,  .996,  .984}38.50  & \cellcolor[rgb]{ 1,  .984,  .937}36.59  & \cellcolor[rgb]{ 1,  .996,  .976}24.42  \\
    PMR   & \cellcolor[rgb]{ 1,  .957,  .929}36.75  & \cellcolor[rgb]{ 1,  .949,  .914}35.71  & \cellcolor[rgb]{ 1,  .949,  .918}20.32  & \cellcolor[rgb]{ 1,  .996,  .976}39.34  & \cellcolor[rgb]{ 1,  .98,  .925}36.97  & \cellcolor[rgb]{ 1,  .996,  .973}25.10  \\
    UME   & 34.91  & 33.41  & \cellcolor[rgb]{ 1,  .894,  .831}21.93  & \cellcolor[rgb]{ 1,  .996,  .973}40.02  & \cellcolor[rgb]{ 1,  .98,  .922}37.12  & \cellcolor[rgb]{ 1,  .988,  .941}30.45  \\
    UMT   & \cellcolor[rgb]{ 1,  .957,  .929}36.72  & \cellcolor[rgb]{ 1,  .973,  .957}34.64  & \cellcolor[rgb]{ 1,  .902,  .839}21.76  & \cellcolor[rgb]{ 1,  .988,  .949}42.58  & \cellcolor[rgb]{ 1,  .992,  .969}35.65  & \cellcolor[rgb]{ 1,  .984,  .929}32.41  \\
    CMCL  & \cellcolor[rgb]{ 1,  .871,  .792}40.21  & \cellcolor[rgb]{ 1,  .89,  .824}38.15  & \cellcolor[rgb]{ 1,  .843,  .749}23.41  & \cellcolor[rgb]{ 1,  .965,  .863}53.31  & \cellcolor[rgb]{ 1,  .973,  .878}38.31  & \cellcolor[rgb]{ 1,  .969,  .878}41.25  \\
    HSR   & \cellcolor[rgb]{ 1,  .839,  .741}41.49  & \cellcolor[rgb]{ 1,  .867,  .78}39.21  & \cellcolor[rgb]{ 1,  .824,  .718}24.01  & \cellcolor[rgb]{ 1,  .961,  .847}55.22  & \cellcolor[rgb]{ 1,  .965,  .847}39.20  & \cellcolor[rgb]{ 1,  .965,  .859}44.67  \\\midrule
    \textbf{Ours} & \cellcolor[rgb]{ .996,  .78,  .647}\textbf{43.85 } & \cellcolor[rgb]{ .996,  .78,  .647}\textbf{42.71 } & \cellcolor[rgb]{ .996,  .78,  .647}\textbf{25.22 } & \cellcolor[rgb]{ 1,  .949,  .8}\textbf{60.52 } & \cellcolor[rgb]{ 1,  .949,  .8}\textbf{40.58 } & \cellcolor[rgb]{ 1,  .949,  .8}\textbf{54.26 } \\
    \bottomrule
    \end{tabular}%
  \label{tab:retrieval}%
\end{table}%

Additionally, we benchmark our approach against recent advancements in retrieval tasks. CMCL\cite{cmcl} introduces a cross-modal center loss for learning distinctive and modality-invariant features, showing impressive results in both in-domain and cross-modal retrieval. HSR\cite{hsr} develops a hierarchical representation strategy, utilizing hierarchical similarity for retrieval tasks.

These comparisons reveal that the challenge of modality competition persists in retrieval tasks. However, ReconBoost effectively mitigates this issue, leading to superior performance.

\subsection{Robustness Performance}

Our initial learning approach assumes all modalities are of high quality. To assess how our method handles noisy data, we introduce Gaussian noise into different modalities and evaluate the performance using the CREMA-D dataset.

\textbf{Case 1:} In scenarios where $50\%$ of the image data is distorted with Gaussian noise $\epsilon \sim \mathcal{N}(\mu,\sigma^2)$ $(\mu = 0)$, we observe the outcomes across various levels of noise intensity $\sigma$, as detailed in the Tab.\ref{tab:noise_img}.

\begin{table}[]
  \centering
  \renewcommand\arraystretch{1.1}
  \caption{Performance comparisons on the CREMA-D dataset in terms of  Acc(\%) when $50\%$ of the image data is corrupted with Gaussian noise i.e., zero mean with the variance of $\sigma^2$.}
  \vskip 0.1in
    \begin{tabular}{cccccc}
    \toprule
    \textbf{Method} & \textbf{$\sigma^2 = 0.0$ } & \textbf{$\sigma^2 = 0.1$} & \textbf{$\sigma^2 = 0.3$} & \textbf{$\sigma^2 = 0.5$} & \textbf{$\sigma^2 = 1.0$} \\
    \midrule
    VisualNet & 50.14  & 47.42  & 43.71  & 40.30  & 35.35  \\
    Concat Fusion & 59.50  & 58.70  & 58.13  & 57.70  & 57.10  \\
    OGM-GE & \cellcolor[rgb]{ 1,  .988,  .941}65.59  & 64.20  & 62.50  & 61.70  & 60.17  \\
    PMR   & \cellcolor[rgb]{ 1,  .984,  .937}66.10  & \cellcolor[rgb]{ 1,  .996,  .976}65.58  & \cellcolor[rgb]{ 1,  .992,  .973}63.39  & \cellcolor[rgb]{ 1,  .996,  .98}62.10  & \cellcolor[rgb]{ 1,  .988,  .953}61.08  \\
    UME   & \cellcolor[rgb]{ 1,  .98,  .914}68.41  & \cellcolor[rgb]{ 1,  .992,  .957}66.49  & \cellcolor[rgb]{ 1,  .996,  .976}63.28  & \cellcolor[rgb]{ 1,  .996,  .984}62.04  & \cellcolor[rgb]{ 1,  .984,  .933}61.46  \\
    UMT   & \cellcolor[rgb]{ 1,  .973,  .89}70.97  & \cellcolor[rgb]{ 1,  .98,  .914}68.76  & \cellcolor[rgb]{ 1,  .98,  .918}64.92  & \cellcolor[rgb]{ 1,  .984,  .937}63.01  & \cellcolor[rgb]{ 1,  .973,  .894}62.23  \\ \hline
    \textbf{Ours} & \cellcolor[rgb]{ 1,  .949,  .8}\textbf{79.82 } & \cellcolor[rgb]{ 1,  .949,  .8}\textbf{74.75 } & \cellcolor[rgb]{ 1,  .949,  .8}\textbf{68.26 } & \cellcolor[rgb]{ 1,  .949,  .8}\textbf{65.73 } & \cellcolor[rgb]{ 1,  .949,  .8}\textbf{63.95 } \\
    \bottomrule
    \end{tabular}%
  \label{tab:noise_img}%
\end{table}%

\textbf{Case 2:} Similarly, when $50\%$ of the audio data encounters the same type of noise distortion, we document the performance changes with different noise intensities $\sigma$, as shown in Tab.\ref{tab:noise_audio}.

\begin{table}[]
  \centering
  \renewcommand\arraystretch{1.1}
  \caption{Performance comparisons on the CREMA-D dataset in terms of Acc(\%) when $50\%$ of the audio data is corrupted with Gaussian noise i.e., zero mean with the variance of $\sigma^2$.}
  \vskip 0.1in
    \begin{tabular}{cccccc}
    \toprule
    \textbf{Method} & \textbf{$\sigma^2 = 0.0$ } & \textbf{$\sigma^2 = 0.1$} & \textbf{$\sigma^2 = 0.3$} & \textbf{$\sigma^2 = 0.5$} & \textbf{$\sigma^2 = 1.0$} \\
    \midrule
    AudioNet & 56.67  & 51.70  & 49.20  & 46.70  & 44.80  \\
    Concat Fusion & \cellcolor[rgb]{ .973,  .984,  .992}59.50  & 57.01  & 55.56  & 54.74  & 52.27  \\
    OGM-GE & \cellcolor[rgb]{ .906,  .945,  .976}65.59  & \cellcolor[rgb]{ .969,  .984,  .992}63.28  & \cellcolor[rgb]{ .957,  .976,  .988}62.09  & \cellcolor[rgb]{ .973,  .984,  .992}59.56  & \cellcolor[rgb]{ .973,  .984,  .992}56.49  \\
    PMR   & \cellcolor[rgb]{ .898,  .941,  .976}66.10  & \cellcolor[rgb]{ .957,  .976,  .988}63.74  & \cellcolor[rgb]{ .933,  .961,  .984}62.83  & \cellcolor[rgb]{ .945,  .969,  .988}60.29  & \cellcolor[rgb]{ .953,  .973,  .988}57.10  \\
    UME   & \cellcolor[rgb]{ .875,  .925,  .969}68.41  & \cellcolor[rgb]{ .973,  .984,  .992}63.01  & \cellcolor[rgb]{ .973,  .984,  .992}61.64  & \cellcolor[rgb]{ .922,  .953,  .98}60.83  & \cellcolor[rgb]{ .902,  .941,  .976}58.57  \\
    UMT   & \cellcolor[rgb]{ .843,  .906,  .961}70.97  & \cellcolor[rgb]{ .902,  .941,  .976}66.29  & \cellcolor[rgb]{ .867,  .922,  .965}64.71  & \cellcolor[rgb]{ .812,  .886,  .953}63.40  & \cellcolor[rgb]{ .843,  .906,  .961}60.37  \\ \hline
    \textbf{Ours} & \cellcolor[rgb]{ .741,  .843,  .933}79.82  & \cellcolor[rgb]{ .741,  .843,  .933}73.65  & \cellcolor[rgb]{ .741,  .843,  .933}68.19  & \cellcolor[rgb]{ .741,  .843,  .933}65.05  & \cellcolor[rgb]{ .741,  .843,  .933}63.24  \\
    \bottomrule
    \end{tabular}%
  \label{tab:noise_audio}%
\end{table}%

\textbf{Case 3:} In cases where both audio and image data are $50\%$ corrupted by Gaussian noise $\epsilon \sim \mathcal{N}(\mu,\sigma^2)$ $(\mu = 0)$, the impacts on performance with varying noise levels $\sigma$ are summarized in the Tab.\ref{tab:noise_both}.

\begin{table}[]
  \centering
  \renewcommand\arraystretch{1.1}
  \caption{Performance comparisons on the CREMA-D dataset in terms of  Acc(\%) when $50\%$ of the image data and the audio data are corrupted with Gaussian noise i.e., zero mean with the variance of $\sigma^2$.}
  \vskip 0.1in
    \begin{tabular}{cccccc}
    \toprule
    \textbf{Method} & \textbf{$\sigma^2 = 0.0$ } & \textbf{$\sigma^2 = 0.1$} & \textbf{$\sigma^2 = 0.3$} & \textbf{$\sigma^2 = 0.5$} & \textbf{$\sigma^2 = 1.0$} \\ 
    \midrule
    AudioNet & 56.67  & 51.70  & 49.20  & 46.70  & 44.80  \\
    VisualNet & 50.14  & 47.42  & 43.71  & 40.30  & 35.35  \\
    Concat Fusion & 59.50  & 55.31  & 52.34  & 49.42  & 47.23  \\
    OGM-GE & 65.59  & 60.14  & 57.36  & 54.26  & 50.38  \\
    PMR   & \cellcolor[rgb]{ 1,  .996,  .992}66.10  & \cellcolor[rgb]{ .996,  .961,  .933}62.57  & \cellcolor[rgb]{ 1,  .984,  .976}58.19  & \cellcolor[rgb]{ 1,  .98,  .969}55.14  & \cellcolor[rgb]{ 1,  .976,  .965}51.25  \\
    UME   & \cellcolor[rgb]{ .996,  .961,  .937}68.41  & \cellcolor[rgb]{ .996,  .953,  .925}62.95  & \cellcolor[rgb]{ .996,  .973,  .953}58.84  & \cellcolor[rgb]{ .996,  .969,  .949}55.72  & \cellcolor[rgb]{ .996,  .957,  .933}51.91  \\
    UMT   & \cellcolor[rgb]{ .992,  .925,  .878}70.97  & \cellcolor[rgb]{ .992,  .918,  .867}65.02  & \cellcolor[rgb]{ .984,  .875,  .8}63.50  & \cellcolor[rgb]{ .98,  .855,  .769}60.63  & \cellcolor[rgb]{ .98,  .851,  .765}55.74  \\ \hline
    \textbf{Ours} & \cellcolor[rgb]{ .973,  .796,  .678}\textbf{79.82 } & \cellcolor[rgb]{ .973,  .796,  .678}\textbf{71.83 } & \cellcolor[rgb]{ .973,  .796,  .678}\textbf{67.17 } & \cellcolor[rgb]{ .973,  .796,  .678}\textbf{63.09 } & \cellcolor[rgb]{ .973,  .796,  .678}\textbf{57.60 } \\
    \bottomrule
    \end{tabular}%
  \label{tab:noise_both}%
\end{table}%

Our observations indicate that despite the presence of noise, our method consistently outperforms competing approaches in all the scenarios mentioned above.


\subsection{Modality Competition Analysis}
\label{app:modality_competition}

In this subsection, we quantify the modality competition and analyze this phenomenon in more detail.

Given input data $\mathcal{X}$ that consists of $M$ modalities,  $\mathcal{X} = \{\mathcal{X}_1, \cdots,  \mathcal{X}_M\}$, $\mathcal{Y}$ represents the ground-truth labels. We can train $M$ separate encoders $\{\varphi_1^{uni}, \cdots, \varphi_{M}^{uni}\}$ and classifiers $\{f_1^{uni}, \cdots,  f_M^{uni}\}$ for each modality through uni-modal training. We can also train encoders $\{\varphi_1^{mul}, \cdots, \varphi_{M}^{mul}\}$ for all modalities through multi-modal learning. Then,  we build a classifier on the frozen modality-specific encoder,  denoted as $\{f_1^{mul}, \cdots,  f_M^{mul}\}$ for all modalities. $\texttt{Acc}(\cdot)$ represents the accuracy evaluation function.

 For any two modalities $\mathcal{X}_i$ (the strong modality) and $\mathcal{X}_j$ (the weak modality), we define the \textbf{modality imbalance ratio (MIR)} in a uni-modal setting as:
\begin{align}
    \text{MIR}^{uni}(\mathcal{X}_i,\mathcal{X}_j) = \frac{\texttt{Acc}(f_i^{uni}\circ \varphi_i^{uni}(\mathcal{X}_i))}{\texttt{Acc}(f_j^{uni}\circ \varphi_j^{uni}(\mathcal{X}_j))}
\end{align}
In a multi-modal setting, the definition of MIR is:
\begin{align}
    \text{MIR}^{mul}(\mathcal{X}_i,\mathcal{X}_j) = \frac{\texttt{Acc}(f_i^{mul}\circ \varphi_i^{mul}(\mathcal{X}_i))}{\texttt{Acc}(f_j^{mul}\circ \varphi_j^{mul}(\mathcal{X}_j))}
\end{align}  

MIR effectively measures the accuracy ratio between any two modalities, where a higher MIR indicates a more pronounced imbalance in learning across different modalities.

Furthermore, to assess the competition between multi-modal and uni-modal learning, we introduce the \textbf{Degree of Modality Competition (DMC)}. Specifically,  DMC compares the MIR of a multi-modal learner to that of a uni-modal learner:

\begin{align}
    \text{DMC}(\mathcal{X}_i,\mathcal{X}_j) = \frac{\text{MIR}^{mul}(\mathcal{X}_i,\mathcal{X}_j)}{\text{MIR}^{uni}(\mathcal{X}_i,\mathcal{X}_j)}
\end{align}

A higher DMC value indicates more intense modality competition. We also expand DMC to accommodate three modalities by calculating the geometric mean of all modality pairs:

\begin{align}
    \text{DMC}(\mathcal{X}_i,\mathcal{X}_j,\mathcal{X}_k) = \sqrt[3]{\prod_{\substack{m,n \in \{i,j,k\}\\ m \neq n}} \text{DMC}(\mathcal{X}_m,\mathcal{X}_n)}
\end{align}

Tab.\ref{tab:mir_dmc} summarizes the modality imbalance ratio of different multi-modal learning methods on both the CREMA-D and AVE datasets. Tab.\ref{tab:rela} shows the DMC value of the concatenation fusion method across all datasets. Notably, as the degree of modality competition rises, so does the improvement our method offers.

\begin{table}[h]
\renewcommand\arraystretch{1.2}
\center
\caption{Modality imbalance ratio (MIR) and the degree of modality competition (DMC) for all competitors on the CREMA-D and AVE dataset. Audio modality is a strong modality.}
\vskip 0.1in
\begin{tabular}{ccccccccc}
    \toprule
    \multirow{2}[4]{*}{\textbf{Method}} & \multicolumn{4}{c}{\textbf{CREMAD Dataset}} & \multicolumn{4}{c}{\textbf{AVE Dataset}} \\
\cmidrule{2-9}          & Audio & Visual & MIR   & DMC   & Audio & Visual & MIR   & DMC \\
    \midrule
    Uni-train & \cellcolor[rgb]{ .898,  .937,  .973}56.67  & \cellcolor[rgb]{ .859,  .914,  .965}50.14  & \cellcolor[rgb]{ .796,  .875,  .945}1.13  & -     & \cellcolor[rgb]{ .984,  .863,  .784}59.37  & \cellcolor[rgb]{ .988,  .914,  .863}30.46  & \cellcolor[rgb]{ .984,  .882,  .816}1.95  & - \\
    Concat Fusion & \cellcolor[rgb]{ .973,  .984,  .992}54.86  & \cellcolor[rgb]{ .973,  .984,  .992}26.81  & \cellcolor[rgb]{ .973,  .984,  .992}2.05  & \cellcolor[rgb]{ .973,  .984,  .992}1.81  & 55.47  & 23.96  & \cellcolor[rgb]{ .996,  .969,  .949}2.32  & \cellcolor[rgb]{ .996,  .969,  .949}1.19  \\
    G-Blending & \cellcolor[rgb]{ .973,  .984,  .992}54.90  & \cellcolor[rgb]{ .969,  .984,  .992}28.05  & \cellcolor[rgb]{ .953,  .973,  .984}1.96  & \cellcolor[rgb]{ .953,  .973,  .984}1.73  & \cellcolor[rgb]{ 1,  .992,  .984}55.80  & 24.12  & \cellcolor[rgb]{ .992,  .965,  .945}2.31  & \cellcolor[rgb]{ .992,  .965,  .945}1.19  \\
    OGM-GE & \cellcolor[rgb]{ .949,  .973,  .988}55.42  & \cellcolor[rgb]{ .961,  .98,  .992}29.17  & \cellcolor[rgb]{ .941,  .965,  .984}1.90  & \cellcolor[rgb]{ .941,  .965,  .984}1.68  & \cellcolor[rgb]{ .996,  .965,  .945}56.51  & \cellcolor[rgb]{ 1,  .98,  .969}25.52  & \cellcolor[rgb]{ .992,  .945,  .91}2.21  & \cellcolor[rgb]{ .992,  .945,  .91}1.14  \\
    PMR   & \cellcolor[rgb]{ .941,  .969,  .984}55.60  & \cellcolor[rgb]{ .961,  .98,  .992}29.21  & \cellcolor[rgb]{ .945,  .965,  .984}1.90  & \cellcolor[rgb]{ .945,  .965,  .984}1.68  & \cellcolor[rgb]{ .992,  .941,  .906}57.20  & \cellcolor[rgb]{ .996,  .969,  .953}26.30  & \cellcolor[rgb]{ .988,  .933,  .898}2.17  & \cellcolor[rgb]{ .988,  .933,  .898}1.12  \\
    UMT   & \cellcolor[rgb]{ .82,  .89,  .953}58.47  & \cellcolor[rgb]{ .878,  .929,  .969}45.69  & \cellcolor[rgb]{ .824,  .894,  .953}1.28  & \cellcolor[rgb]{ .824,  .894,  .953}1.13  & \cellcolor[rgb]{ .976,  .816,  .71}60.70  & \cellcolor[rgb]{ .988,  .906,  .851}31.07  & \cellcolor[rgb]{ .984,  .882,  .816}1.95  & \cellcolor[rgb]{ .984,  .882,  .816}1.00  \\\midrule
    \textbf{Ours}  & \cellcolor[rgb]{ .741,  .843,  .933}\textbf{60.23 } & \cellcolor[rgb]{ .741,  .843,  .933}\textbf{73.01 } & \cellcolor[rgb]{ .741,  .843,  .933}\textbf{0.82 } & \cellcolor[rgb]{ .741,  .843,  .933}\textbf{0.73 } & \cellcolor[rgb]{ .973,  .796,  .678}\textbf{61.20 } & \cellcolor[rgb]{ .973,  .796,  .678}\textbf{39.06 } & \cellcolor[rgb]{ .973,  .796,  .678}\textbf{1.57 } & \cellcolor[rgb]{ .973,  .796,  .678}\textbf{0.80 } \\
    \bottomrule
    \end{tabular}%
\label{tab:mir_dmc}
\end{table}

\begin{table*}[h]
\renewcommand\arraystretch{1.2}
\center
\caption{The correlation between the Degree of Modality Competition (DMC) using the concatenation fusion method and the enhancement of our method compared to that across all datasets. If the dataset lacks this modality, it is denoted as '-'.}
\vskip 0.1in
\begin{tabular}{ccccccccccc}
\toprule
\multirow{2}{*}{Dataset} & \multicolumn{3}{c}{Uni-modal} & \multicolumn{3}{c}{Concat-fusion } & \multirow{2}{*}{DMC} & \multirow{2}{*}{Concat} & \multirow{2}{*}{Ours} & \multirow{2}{*}{Relative Improvement} \\ \cline{2-7}
 & Audio & Visual & Text & Audio & Visual & Text &  &  &  &  \\ \hline
CREMA-D & 56.67 & 50.14 & - & 54.86 & 26.81 & - & \textbf{1.81} & 59.50 & 79.82 & \textbf{34.15\%} \\ 
AVE & 59.37 & 30.46 & - & 55.47 & 23.96 & - & \textbf{1.19} & 62.68 & 71.35 & \textbf{13.83\%} \\
MOSEI & 52.29 & 50.35 & 66.41 & 49.02 & 49.02 & 66.13 & \textbf{1.01} & 66.71 & 68.61 & \textbf{2.85\%} \\
MOSI & 54.81 & 57.87 & 75.94 & 54.25 & 54.37 & 74.05 & \textbf{0.99} & 76.23 & 77.96 & \textbf{2.27\%}\\ 
CH-SIMS & 58.20 & 63.02 & 70.45 & 54.27 & 59.74 & 68.71 & \textbf{1.03} & 71.55 & 73.88 & \textbf{3.26\%} \\  \bottomrule
\end{tabular}
\label{tab:rela}
\end{table*}

\subsection{Analysis of Mutual Information in Modalities}
\label{sec:app_MI}
\begin{figure*}[]  
\centering  
\includegraphics[width=0.5\linewidth]{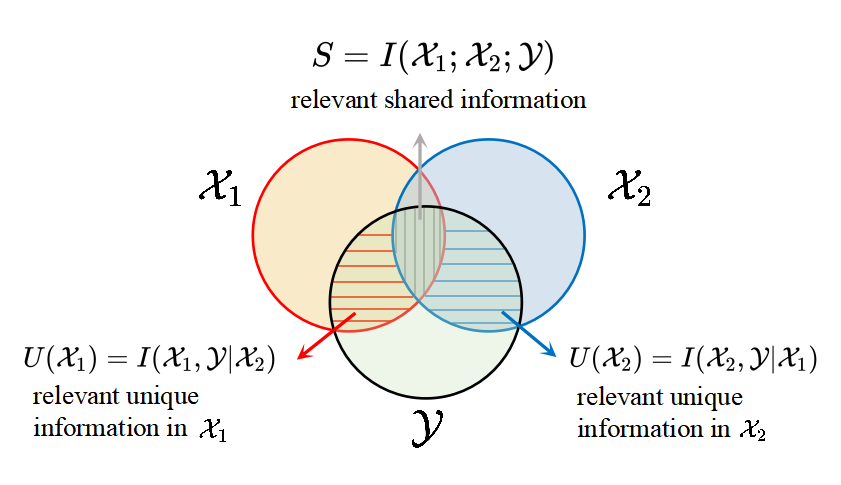}
\caption{The visualization of the multi-modal information. This figure is derived from \cite{factor-contrastive-learning}.}  
\label{fig:visual_mi} 
\end{figure*}

In this subsection, we will illustrate the effectiveness of our method from the perspective of mutual information. Assume that two modalities are denoted as $\mathcal{X}_1$ and $\mathcal{X}_2$. $\mathcal{Y}$ represents the groud-truth labels. Following \cite{factor-contrastive-learning}, we decompose the multi-modal information $I(\mathcal{X}_1,\mathcal{X}_2;\mathcal{Y})$ into three conditional mutual information (MI) terms and visualize the multi-modal information as Fig.\ref{fig:visual_mi}.
\begin{align}
   I(\mathcal{X}_1,\mathcal{X}_2;\mathcal{Y}) = \underbrace{I(\mathcal{X}_1;\mathcal{X}_2;\mathcal{Y})}_{S(\mathcal{X}_1,\mathcal{X}_2) = \text{relevant shared info.}} + \underbrace{I(\mathcal{X}_1,\mathcal{Y}|\mathcal{X}_2)}_{U(\mathcal{X}_1) = \text{relevant unique info. in }\mathcal{X}_1} + \underbrace{I(\mathcal{X}_2,\mathcal{Y}|\mathcal{X}_1)}_{U(\mathcal{X}_2) = \text{relevant unique info. in }\mathcal{X}_2}
\end{align}

When only one modality $\mathcal{X}_1$ exists, the uni-modal only contains unique information $U(\mathcal{X}_1)$. Then, in a multi-modal setting, maximize the information that $\mathcal{X}_2$ can bring becomes the key to improving the performance of multi-modal learning algorithms. We measure the information that $\mathcal{X}_2$ can bring using the difference in accuracy between using the multi-modal approach and the uni-modal model, as:
\begin{align}
    U(\mathcal{X}_2) + S(\mathcal{X}_1,\mathcal{X}_2) = \texttt{Acc}(\mathcal{X}_1,\mathcal{X}_2) - \texttt{Acc}(\mathcal{X}_1)
\end{align}
where $\texttt{Acc}(\mathcal{X}_1)$ and $\texttt{Acc}(\mathcal{X}_2)$ denote the accuracy of the unimodal learning algorithm using only $\mathcal{X}_1$ and $\mathcal{X}_2$ modalities, respectively; $\texttt{Acc}(\mathcal{X}_1,\mathcal{X}_2)$ denote the accuracy of the multi-modal learning algorithm using both $\mathcal{X}_1$ and $\mathcal{X}_2$ modalities.

\begin{figure}[]
\centering  
\subfigure[Audio on CREMA-D]{
\includegraphics[width=0.31\columnwidth]{figure/mi_cremad_audio.pdf}}
\subfigure[Visual on CREMA-D]{
\includegraphics[width=0.31\columnwidth]{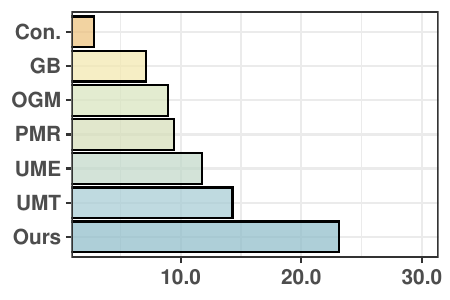}}
\subfigure[View on MN40]{
\includegraphics[width=0.31\columnwidth]{figure/mi_mn40_view.pdf}}
\subfigure[Audio on AVE]{
\includegraphics[width=0.33\columnwidth]{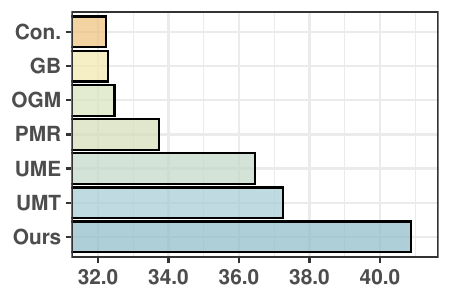}}
\subfigure[Visual on AVE]{
\includegraphics[width=0.33\columnwidth]{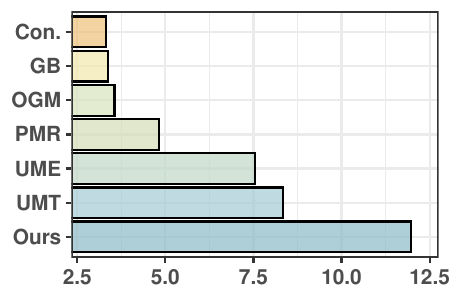}}
\caption{Quantitative analysis of task relevant mutual information in modalities on the AVE, CREMA-D, and MN40 datasets.}
\label{fig:MI}
\end{figure}

Then, we evaluate it among different competitors on three benchmarks. Overall, as shown in Fig.\ref{fig:MI}, our method consistently outperforms others, demonstrating the potential of maximizing the valuable information in each modality. This further illustrates the effectiveness of our method.

\subsection{Modality Selection Strategy}

In this subsection, we investigate the effect of different modality selection strategies. Our method expects to tackle the issue of modality competition. To this end, we alternate learning for each modality. This approach intuitively eases modality competition in gradient space by requiring separate use of modality-specific gradients. We now explore quality-guided criteria for modality selection. We introduce two additional modality selection schemes based on loss value as a measure of modality-specific data quality:
\begin{itemize}
    \item $S_1$:  We select the modality learner with the lowest loss value for updates in each round, favoring high-quality modalities.
    \item $S_2$: We select the modality learner with the highest loss value for updates in each round, prioritizing low-quality modalities.
\end{itemize}
We evaluate the effectiveness of these two optimization orders using the AVE dataset, as shown in Tab.\ref{tab:selection}.

The results demonstrate that our method surpasses those based on quality selection. This might be because the $S_1$ strategy could cause low-quality modality learners to get stuck at poor local optima. Conversely, the $S_2$ strategy may restrict the potential of high-quality modalities.

\begin{table*}[h]
\renewcommand\arraystretch{1.2}
\center
\caption{Performance comparisons on the AVE dataset in terms of Acc(\%) with different selection strategies.}
\vskip 0.1in
\begin{tabular}{c|ccc}
    \toprule
    Optimization Order & Overall Acc & Audio Acc & Visual Acc \\
    \midrule
    $S_1$  & \cellcolor[rgb]{ .949,  .969,  .988}61.45  & \cellcolor[rgb]{ .804,  .882,  .953}60.03  & \cellcolor[rgb]{ .949,  .969,  .988}24.11  \\
    $S_2$  & \cellcolor[rgb]{ .796,  .878,  .953}68.75  & \cellcolor[rgb]{ .949,  .969,  .988}57.20  & \cellcolor[rgb]{ .745,  .847,  .941}38.90  \\
    \midrule
    \textbf{Ours} & \cellcolor[rgb]{ .741,  .843,  .937}\textbf{71.35 } & \cellcolor[rgb]{ .741,  .843,  .937}\textbf{61.20 } & \cellcolor[rgb]{ .741,  .843,  .937}\textbf{39.06 } \\
    \bottomrule
    \end{tabular}%
    \label{tab:selection}
\end{table*}

\subsection{Applicable to Other Fusion Schemes}
\label{sec:app_fusion}
In this subsection, we investigate how to combine our approach with some multi-modal fusion methods to better improve performance. Multi-modal fusion methods are typically divided into two categories: feature-level and decision-level fusions \cite{mm-survey}. Feature-level fusion combines latent features before making predictions, commonly used in multi-modal joint-learning approaches. In contrast, decision-level fusion aggregates predictions from each modality to reach a final decision. Our main paper demonstrates that joint learning can cause modality competition. To mitigate this, we introduced a new multi-modal learning framework based on decision-level fusion strategies, enhancing adaptability to complex decision-making scenarios. We modify our original decision aggregation formula as follows:
\begin{align}
    \Phi_{M}(x_i) = \sum_{k=1}^{M}w_k\cdot\phi_k(\vartheta_k;m_i^k)
\end{align}
where $w_k$ signifies the importance of the $k$-th modality during inference.

Our ReconBoost framework can be easily combined with other fusion methods, particularly:
\begin{itemize}
    \item \textbf{QMF} \cite{dynamic_fusion} employs a dynamic, uncertainty-aware weighting mechanism at the decision level.
    \item \textbf{TMC} \cite{TMC} uses a dynamic approach to integrate modalities through the Dempster-Shafer theory efficiently.
\end{itemize}
Additionally, we benchmark against two straightforward baselines:
\begin{itemize}
    \item Learnable Weighting (\textbf{LW}): Assigns trainable weights $w_k$ to each modality and learns these weights alongside other parameters.
    \item Naive Averaging (\textbf{NA}): Averages predictions across modalities, setting $w_k=1$ for all modalities.
\end{itemize}

Furthermore, to emphasize the superiority of our approach, we also evaluate a novel feature-based fusion competitor, MMTM \cite{mmtm}, on the AVE and CREMA-D datasets.

\begin{table}[htbp]
\renewcommand\arraystretch{1.2}
  \centering
  \caption{Performance comparisons on the AVE and CREMA-D dataset in terms of Acc(\%) with different fusion strategies.}
  \vskip 0.1in
    \begin{tabular}{c|cccccc}
    \toprule
    \multicolumn{1}{c}{\multirow{2}[4]{*}{\textbf{Method}}} & \multicolumn{3}{c}{\textbf{AVE}} & \multicolumn{3}{c}{\textbf{CREMA-D}} \\
\cmidrule{2-7}    \multicolumn{1}{c}{} & Overall Acc & Audio Acc & Visual Acc & Overall Acc & Audio Acc & Visual Acc \\
    \midrule
    OGM\_GE + MMTM & \cellcolor[rgb]{ .973,  .984,  .992}66.14  & \cellcolor[rgb]{ .973,  .984,  .992}58.23  & \cellcolor[rgb]{ .973,  .984,  .992}28.09  & 69.83 & 58.76 & 53.35 \\
    PMR + MMTM & \cellcolor[rgb]{ .922,  .953,  .98}67.72  & \cellcolor[rgb]{ .961,  .976,  .992}58.47  & \cellcolor[rgb]{ .965,  .98,  .992}28.73  & \cellcolor[rgb]{ .996,  1,  .992}70.14 & \cellcolor[rgb]{ .984,  .992,  .976}58.94 & \cellcolor[rgb]{ .992,  .996,  .988}54.23 \\
    UMT + MMTM & \cellcolor[rgb]{ .843,  .906,  .961}70.16  & \cellcolor[rgb]{ .839,  .906,  .961}60.40  & \cellcolor[rgb]{ .835,  .902,  .957}35.83  & \cellcolor[rgb]{ .914,  .953,  .882}74.35 & \cellcolor[rgb]{ .788,  .886,  .718}60.86 & \cellcolor[rgb]{ .898,  .945,  .867}62.83 \\\midrule
    Ours + NA & \cellcolor[rgb]{ .804,  .882,  .949}71.35  & \cellcolor[rgb]{ .792,  .875,  .949}61.20  & \cellcolor[rgb]{ .776,  .867,  .945}39.06  & \cellcolor[rgb]{ .804,  .894,  .741}79.82 & \cellcolor[rgb]{ .851,  .922,  .804}60.23 & \cellcolor[rgb]{ .788,  .886,  .722}73.01 \\
    Ours + LW & \cellcolor[rgb]{ .769,  .863,  .941}72.40  & \cellcolor[rgb]{ .784,  .871,  .945}61.31  & \cellcolor[rgb]{ .773,  .863,  .945}39.13  & \cellcolor[rgb]{ .8,  .89,  .733}80.11 & \cellcolor[rgb]{ .867,  .929,  .824}60.09 & \cellcolor[rgb]{ .784,  .882,  .718}73.30 \\
    Ours + TMC & \cellcolor[rgb]{ .753,  .851,  .937}72.96  & \cellcolor[rgb]{ .773,  .863,  .941}61.51  & \cellcolor[rgb]{ .757,  .851,  .937}40.20  & \cellcolor[rgb]{ .788,  .886,  .718}80.68 & \cellcolor[rgb]{ .835,  .914,  .784}60.37 & \cellcolor[rgb]{ .78,  .882,  .71}73.86 \\
    Ours + QMF & \cellcolor[rgb]{ .741,  .843,  .933}\textbf{73.20}  & \cellcolor[rgb]{ .741,  .843,  .933}\textbf{61.96}  & \cellcolor[rgb]{ .741,  .843,  .933}\textbf{40.85}  & \cellcolor[rgb]{ .776,  .878,  .706}\textbf{81.11} & \cellcolor[rgb]{ .776,  .878,  .706}\textbf{60.94} & \cellcolor[rgb]{ .776,  .878,  .706}\textbf{73.87} \\
    \bottomrule
    \end{tabular}%
  \label{tab:fusion}%
\end{table}%

Overall, as shown in Tab.\ref{tab:fusion}, our method consistently outperforms others, highlighting the potential of more flexible fusion strategies to enhance performance. This reaffirms the effectiveness of ReconBoost.

\subsection{Impact of Different Classifiers}

In Tab.\ref{tab:uni}, we limit classifiers to linear models to assess the effectiveness of our proposed ensemble method. Herein, we expand our evaluation to include non-linear classifiers featuring multiple fully connected (FC) layers and ReLU functions. Specifically, we develop non-linear classifier architecture, Fc+Relu+Fc, and FC+Relu+FC+Relu+FC, for the encoders used in all methods and test its performance on the CREMA-D dataset. As shown in Tab.\ref{tab:clf}, we arrive at the conclusion that $1)$ modality competition exists no matter which classifier is used. $2)$ Our approach, ReconBoost, enhances the performance of encoders with various classifiers, demonstrating that our model effectively learns high-quality latent features.

\begin{table}[]
\renewcommand\arraystretch{1.2}
  \centering
  \caption{Performance comparisons on the CREMA-D dataset in terms of Acc(\%) with different classifiers.}
   \vskip 0.1in
    \begin{tabular}{ccccccc}
    \toprule
    \multirow{2}[4]{*}{\textbf{Method}} & \multicolumn{2}{c}{\textbf{FC}} & \multicolumn{2}{c}{\textbf{FC+Relu+FC}} & \multicolumn{2}{c}{\textbf{FC+Relu+FC+Relu+FC}} \\
\cmidrule{2-7}          & Visual &  Audio & Visual & Audio & Visual & Audio \\
    \midrule
    Uni-train & \cellcolor[rgb]{ .89,  .941,  .855}50.14 & \cellcolor[rgb]{ .925,  .961,  .902}56.67 & \cellcolor[rgb]{ .89,  .941,  .855}50.25 & \cellcolor[rgb]{ .914,  .957,  .89}56.97 & \cellcolor[rgb]{ .89,  .941,  .855}50.31 & \cellcolor[rgb]{ .914,  .953,  .886}57.10 \\
    Concat Fusion & 26.81 & 54.86 & 26.89 & 54.91 & 26.96 & 55.02 \\
    OGM-GE & \cellcolor[rgb]{ .992,  .996,  .988}29.17 & \cellcolor[rgb]{ .98,  .988,  .973}55.42 & \cellcolor[rgb]{ .988,  .996,  .984}29.72 & \cellcolor[rgb]{ .957,  .976,  .941}56.03 & \cellcolor[rgb]{ .988,  .996,  .984}29.91 & \cellcolor[rgb]{ .957,  .976,  .941}56.11 \\
    PMR   & \cellcolor[rgb]{ .992,  .996,  .988}29.21 & \cellcolor[rgb]{ .973,  .984,  .961}55.60 & \cellcolor[rgb]{ .988,  .996,  .984}29.81 & \cellcolor[rgb]{ .945,  .973,  .929}56.22 & \cellcolor[rgb]{ .988,  .992,  .984}30.05 & \cellcolor[rgb]{ .941,  .969,  .922}56.45 \\
    UMT   & \cellcolor[rgb]{ .91,  .953,  .882}45.69 & \cellcolor[rgb]{ .851,  .922,  .804}58.47 & \cellcolor[rgb]{ .906,  .949,  .875}46.73 & \cellcolor[rgb]{ .843,  .914,  .792}58.71 & \cellcolor[rgb]{ .906,  .949,  .875}47.01 & \cellcolor[rgb]{ .839,  .914,  .788}58.93 \\
    \textbf{Ours} & \cellcolor[rgb]{ .776,  .878,  .706}\textbf{73.01} & \cellcolor[rgb]{ .776,  .878,  .706}\textbf{60.23} & \cellcolor[rgb]{ .776,  .878,  .706}\textbf{73.34} & \cellcolor[rgb]{ .776,  .878,  .706}\textbf{60.24} & \cellcolor[rgb]{ .776,  .878,  .706}\textbf{73.86} & \cellcolor[rgb]{ .776,  .878,  .706}\textbf{60.37} \\
    \bottomrule
    \end{tabular}%
  \label{tab:clf}%
\end{table}%

\subsection{Sensitivity Analysis of $\lambda$}
\label{app:lambda}
To assess the impact of $\lambda$, we carry out additional sensitivity tests by altering $\lambda$'s value. We present the results for CREMA-D, AVE, and ModelNet40 in the Tab.\ref{tab:lambda}. The performance of our method stays consistent with $\lambda$ values between $[1/4, 1/2]$. Additionally, our method continues to achieve state-of-the-art results within this $\lambda$ range.

\begin{table*}[htbp]
\renewcommand\arraystretch{1.2}
  \centering
  \caption{Sensitivity analysis of $\lambda$ on the CREMA-D, AVE and ModelNet40 datasets.}
  \vskip 0.1in
    \begin{tabular}{c|ccc}
    \toprule
    \multicolumn{1}{c}{\textbf{Method}} & \textbf{AVE} & \textbf{CREMA-D} & \textbf{MN40} \\
    \midrule
    Concat Fusion & 62.68 & 59.50  & 83.18 \\
    G-Blending  & 62.75 & \cellcolor[rgb]{ .996,  .957,  .933}63.81 & \cellcolor[rgb]{ .996,  .969,  .949}84.56 \\
    OGM-GE & \cellcolor[rgb]{ 1,  .996,  .992}62.93 & \cellcolor[rgb]{ .992,  .941,  .906}65.59 & \cellcolor[rgb]{ .996,  .945,  .91}85.61 \\
    PMR   & \cellcolor[rgb]{ .996,  .965,  .945}64.20 & \cellcolor[rgb]{ .992,  .937,  .898}66.10 & \cellcolor[rgb]{ .992,  .929,  .89}86.20 \\
    UME   & \cellcolor[rgb]{ .988,  .902,  .843}66.92 & \cellcolor[rgb]{ .988,  .914,  .863}68.41 & \cellcolor[rgb]{ .996,  .949,  .922}85.37 \\
    UMT   & \cellcolor[rgb]{ .984,  .882,  .816}67.71 & \cellcolor[rgb]{ .988,  .886,  .82}70.97 & \cellcolor[rgb]{ .98,  .839,  .745}90.07 \\\midrule
    Ours $\lambda = 1/4$ & \cellcolor[rgb]{ .973,  .796,  .678}\textbf{71.35} & \cellcolor[rgb]{ .976,  .804,  .69}79.26 & \cellcolor[rgb]{ .976,  .812,  .706}91.13 \\
    Ours $\lambda = 1/3$ & \cellcolor[rgb]{ .976,  .824,  .718}70.31 & \cellcolor[rgb]{ .973,  .796,  .678}\textbf{79.82} & \cellcolor[rgb]{ .973,  .796,  .678}\textbf{91.78} \\
    Ours $\lambda = 1/2$ & \cellcolor[rgb]{ .98,  .839,  .749}69.53 & \cellcolor[rgb]{ .976,  .824,  .722}77.13 & \cellcolor[rgb]{ .976,  .812,  .702}91.25 \\
    Ours $\lambda = 1$ & \cellcolor[rgb]{ .98,  .843,  .749}69.49 & \cellcolor[rgb]{ .988,  .894,  .831}70.31 & \cellcolor[rgb]{ .98,  .843,  .749}89.91 \\
    \bottomrule
    \end{tabular}%
  \label{tab:lambda}%
\end{table*}%

\subsection{Latent Embedding Visualization}
\label{app:tsne}
Taking a step further, we visualize the latent embedding of modality-specific features among different competitors on the CREMA-D datasets. Specifically, we first regard the outputs of the backbone as the latent vectors of images and then project them into a 2D case by t-SNE \cite{tsne}. Comparing these results, we can see that our proposed method outperforms other competitors in all modalities since the cluster results of ReconBoost are more significant, especially in the weak modality. This again ascertains the advantages of our proposed approach.

\begin{figure*}[h]  
\centering  
\includegraphics[width=0.99\linewidth]{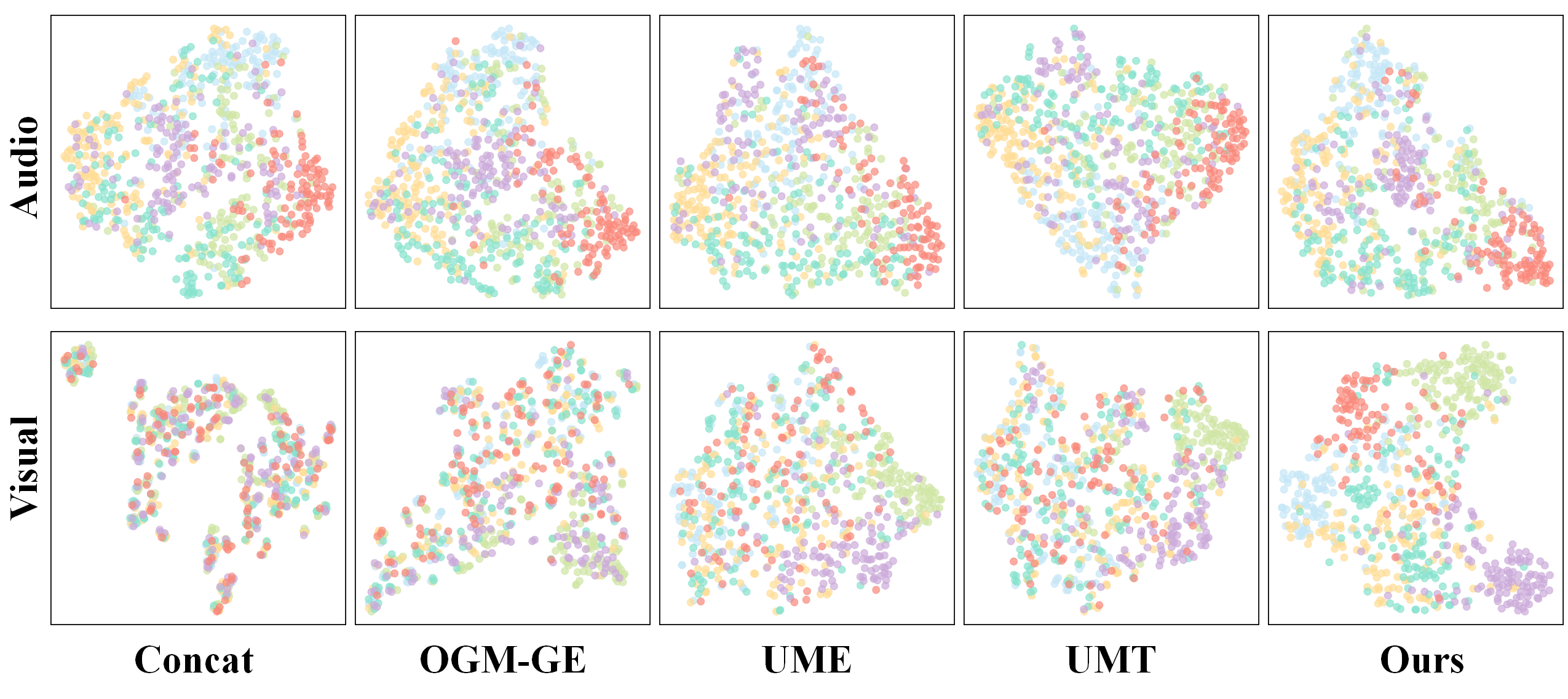}
\caption{The visualization of the modality-specific feature among competitors in the CREMA-D dataset by using the t-SNE method \cite{tsne}.}  
\label{fig:tsne_all} 
\end{figure*}

  

\end{document}